\documentclass[]{style/iromlab}

\usepackage{amsmath, amsfonts, amssymb, amsthm}
\usepackage{mathtools}
\usepackage{bm}
\usepackage{bbm}
\usepackage{units}

\usepackage{booktabs}
\usepackage{multirow}
\usepackage{tabularx}
\usepackage{longtable}
\usepackage{threeparttable}
\usepackage{subcaption}
\usepackage{adjustbox}

\usepackage{graphicx}
\usepackage{wrapfig}
\usepackage{float}
\usepackage{placeins}

\usepackage{enumerate}
\usepackage[inline]{enumitem}

\usepackage[ruled,algo2e]{algorithm2e}
\usepackage{algpseudocode}

\usepackage{xcolor}
\usepackage{tcolorbox}

\usepackage[numbers,sort&compress]{natbib}
\usepackage{cleveref}

\usepackage{multicol}
\usepackage{caption}
\usepackage{balance}
\usepackage{microtype}

\usepackage{pifont}
\usepackage{etoolbox}
\usepackage{xspace}

\usepackage[title]{appendix}

\definecolor{claude_color}{HTML}{F89E62}
\definecolor{deepseek_color}{HTML}{78B6E8}
\definecolor{o3_mini_color}{HTML}{6CD5A1}
\definecolor{red_color}{HTML}{E13B55}
\definecolor{prompt_color}{HTML}{71502B}
\newcommand*{\algname}{\textcolor{black}{\text{PlayWorld}}\xspace}
\definecolor{mygreen}{RGB}{105,192,90}

\tcbset{
  promptbox/.style={
    colback=gray!6,
    colframe=gray!25,
    boxrule=0.4pt,
    arc=2pt,
    left=6pt,
    right=6pt,
    top=4pt,
    bottom=4pt
  }
}

\tcbset{
  promptstyle_custom/.style={
    colback=deepseek_color!10,
    colframe=deepseek_color!60,
    coltitle=black,
    boxrule=1.0pt,
    arc=2pt,
    outer arc=2pt,
    left=4pt,
    right=4pt,
    top=4pt,
    bottom=4pt,
    fonttitle=\bfseries,
  }
}

\makeatletter
\newcommand{\longdash}[1][2em]{%
  \makebox[#1]{$\m@th\smash-\mkern-7mu\cleaders\hbox{$\mkern-2mu\smash-\mkern-2mu$}\hfill\mkern-7mu\smash-$}}
\makeatother

\author[1]{Tenny Yin}
\author[1]{Zhiting Mei}
\author[1]{Zhonghe Zheng}
\author[1]{Miyu Yamane}
\author[1]{David Wang}
\author[1]{Jade Sceats}
\author[1]{\mbox{Samuel M. Bateman}}
\author[1]{Lihan Zha}
\author[1]{Apurva Badithela}
\author[1]{Ola Shorinwa}
\author[1]{Anirudha Majumdar}

\affiliation[1]{Princeton University}

\begin{document}

\title{
\algname: Learning Robot World Models from Autonomous Play
}

\abstract{
Action-conditioned video models offer a promising path to building general-purpose robot simulators that can improve directly from data. 
Yet, a critical question remains: \emph{what data is needed to train them effectively?}
While existing models typically train on human-collected demonstrations or task-specific policy rollouts, we hypothesize that this leads to hallucinations caused by systematic biases and poor coverage of interaction dynamics.  
In contrast, we present \emph{\algname}: a new data paradigm for training robot world models through \emph{autonomous play}.
We demonstrate a pipeline for scalable (including overnight) unsupservised data collection, enabling broad coverage of complex, long-tailed physical interactions essential for modeling realistic object dynamics. 
Experiments across diverse manipulation tasks show that \algname generates high-quality, physically consistent predictions for contact-rich interactions that are not captured by world models trained on human-collected data.
We further demonstrate the versatility of \algname in enabling fine-grained failure prediction and policy evaluation, with up to 40\% improvements over human-collected data. Finally, we demonstrate how \algname enables reinforcement learning in the world model, improving policy performance by 65\% in success rates when deployed in the real world. To our knowledge, \algname represents the first work to demonstrate autonomous play as an effective training paradigm for video world models.
}



\website{
https://robot-playworld.github.io/
}
{
robot-playworld.github.io
}

\maketitle

\section{Introduction}

Generative video models~\cite{agarwal2025cosmos,wan2025wan,deepmind_veo3_techreport, kong2024hunyuanvideo} hold tantalizing potential to serve as general-purpose data-driven simulators for robotics~\cite{huang2025vid2worldcraftingvideodiffusion, deepmind2025genie3, mei2026video}. 
In principle, video models address some of the most critical challenges faced by traditional physics-based simulators by generating highly photorealistic outputs, simulating non-rigid-body interactions (e.g., deformable objects, liquids), providing a direct real-to-sim pipeline by conditioning video generation on images of real-world scenes, and affording the ability to close the sim-to-real gap by scaling data.
An emerging line of work in robotics~\cite{jang2025dreamgen, guo2025ctrl, kim2026cosmos, li2025worldeval, feng2025vidar, liang2025video, veorobotics2025} seeks to reap these benefits by fine-tuning video generation backbones pre-trained on internet-scale data~\cite{blattmann2023stable, agarwal2025cosmos} with robotics datasets, and using the resulting models for scalable data generation~\cite{liang2024dreamitate, patel2025robotic, jang2025dreamgen, guo2026vlawiterativecoimprovementvisionlanguageaction}, policy evaluation~\cite{quevedo2025worldgymworldmodelenvironment, li2025worldeval, guo2025ctrl, veorobotics2025}, and planning~\cite{kim2026cosmos, liang2025video, chen2025largevideoplanner}. 

Despite this progress, state-of-the-art video models remain far from reliable world simulators. 
While capable of generating physically consistent, photorealistic, and long-horizon rollouts in non-interactive scenes (e.g., autonomous driving \cite{hu2023gaia1generativeworldmodel, ren2025gen3c}), they remain vulnerable to hallucinations when simulating contact-rich interactions: from objects that duplicate when grasped, to ones that appear, disappear, or move and deform in unrealistic ways upon contact \cite{denton2018stochastic, liu2024physgen, akkerman2025interdyn, ai2025review, deepmind_veo3_techreport, bansal2024videophy, mei2025confident, mei2025world, veorobotics2025}. 
As a result, the use of video models in robotic manipulation has largely been restricted to tasks with minimal contact and focused around high-level language-following abilities \cite{veorobotics2025, guo2025ctrl}.

In this paper, we take a \emph{data-centric perspective} on training high-quality video models for robotic manipulation. 
Our core hypothesis is that the aforementioned challenges with simulating contact-rich dynamics stem from the data on which video models are trained or fine-tuned. 
In particular, existing action-conditioned video models for manipulation (e.g., ~\cite{quevedo2025worldgymworldmodelenvironment, li2025worldeval, guo2025ctrl}) are \emph{almost exclusively trained on human demonstration datasets}~\cite{o2024open, khazatsky2025droidlargescaleinthewildrobot} developed for imitation learning. 
This inherently limits data coverage to a narrow distribution of states focused along \emph{successful} task executions. 
As a result, this data provides video models with little supervision on complex contact dynamics and state transitions that counterfactual actions might introduce \cite{lu2020sample, pitis2020counterfactual}. 
Small prediction errors at critical contact events can thus compound quickly into qualitatively divergent policy rollouts which are often heavily biased towards successful executions seen in the data (Sec.~\ref{sec:evaluation}).

\begin{figure}
  \includegraphics[width=\linewidth]{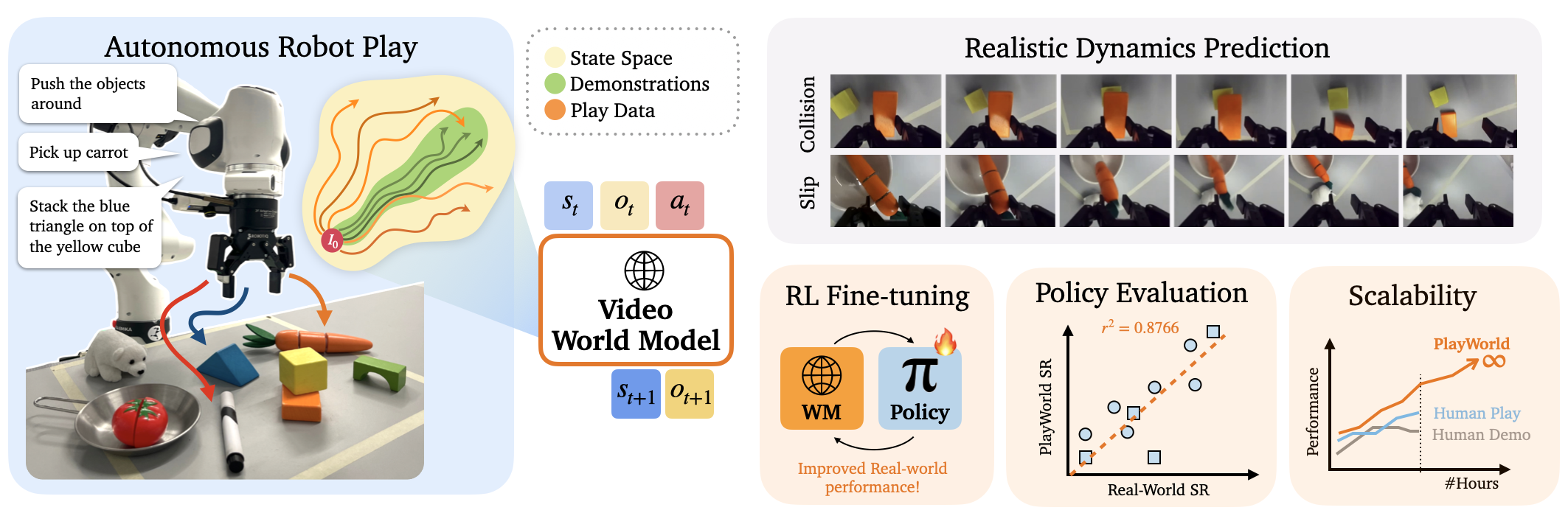}
    \caption{We introduce \algname, a scalable framework for training high-fidelity video world models from autonomous robot play that enables fine-grained dynamics prediction for accurate policy evaluation, and online reinforcement learning policy fine-tuning that yields strong real-world success rate improvements.}
    \label{fig:banner}
\end{figure}

How can we obtain the most useful data for learning high-quality world models?
From a data-centric standpoint, we argue that the right training corpus needs to (i) be generated through \emph{directed exploration} that continually expands coverage of diverse contact events, and (ii) support \emph{efficient scaling}, so exploration can grow naturally with minimal manual engineering or supervision.
Inspired by work in developmental psychology~\cite{Hoch2019Journey, weisberg2016guidedplay}, we ask: can robots actively curate a diverse set of experiences for training high-quality video models through \emph{autonomous play}~\cite{lynch2019learninglatentplansplay, cui2022playpolicyconditionalbehavior, zhou2024autonomousimprovementinstructionfollowing}? 
By interacting with objects in a semi-structured fashion without human supervision, autonomous play data can (i) induce broad coverage of relevant state transitions that might occur during policy execution, and (ii) be highly scalable in terms of both quantity and diversity of objects and scenes. 
These advantages make autonomous play a compelling data-collection framework for contact-rich world modeling. 

{\bf Contributions.} We introduce \emph{\algname}, a simple and scalable data paradigm for training robot world models based on autonomous play, demonstrating that broad, interaction-rich coverage is the key factor for learning accurate, hallucination-free dynamics.
With this framework, we make the following contributions:

\begin{itemize}
    \item[(1)] Through careful empirical analysis, we show that \algname produces substantially \emph{broader coverage} of interaction dynamics, generating significantly more diverse contact events, object states, and failure modes than human-collected data.
    
    \item[(2)] By minimizing human intervention to occasional monitoring and scene resets, we demonstrate that \algname enables highly \emph{scalable} data collection, including fully unsupervised overnight operation.

    \item[(3)] Across a range of manipulation tasks, we show that \algname enables simulation of \emph{realistic physical interactions} with fine-grained predictive accuracy, yielding significant gains on perceptual metrics and continues to improve with \emph{scale}, even beyond $5\times$ the regime where performance saturates with human demonstration data.
    

    \item[(4)] We demonstrate that \algname supports effective post-training applications, including \emph{fine-grained policy evaluation} that strongly correlates with real-world performance, and \emph{closed-loop RL fine-tuning} that achieves up to $65\%$ improvement in real-world success rates over the pre-trained policy.
    
\end{itemize}
To our knowledge, \algname is the \emph{first} work to establish the practicality and advantages of autonomous robot play data for training action-conditioned video models.

\section{Related Work}
\label{sec:related_work}

\subsection{World Models for Robotics}
World models~\cite{ha2018worldmodels, hafner2020dreamcontrollearningbehaviors, hansen2024tdmpc2scalablerobustworld} learn predictive models of environment dynamics, enabling agents to reason and plan beyond purely reactive policies~\cite{hafner2023mastering, zhou2025dinowmworldmodelspretrained, nakamura2025generalizingsafetycollisionavoidancelatentspace}.
A large body of work in model-based reinforcement learning (RL) has demonstrated that large-scale interaction data with broad state-action coverage is necessary for training effective latent-space world models, which is often collected via on-policy or curiosity-driven exploration in simulation~\cite{pathak2017curiosity, sekar2020planning, hafner2023mastering, hansen2023td}. 
However, achieving the same levels of scale via exploration in real-world environments is challenging, posing a major bottleneck to collecting diverse interaction data required for training generalizable models~\cite{wu2023daydreamer}.
As a result, many robotic world models are trained from either manually collected on-policy roll-outs~\cite{jiang2025world4rl, kim2026cosmos, zhu2025wmpo, guo2026vlawiterativecoimprovementvisionlanguageaction} or large offline datasets~\cite{quevedo2025worldgymworldmodelenvironment, guo2025ctrl, veorobotics2025, assran2025v}, which are challenging to scale on hardware and can be biased toward narrow behaviors, limiting their ability to model the dynamical effects of diverse actions \cite{pitis2020counterfactual}.
We address this limitation by designing a system to collect large-scale exploratory interaction data on hardware, expanding coverage over states and actions for learning generalizable dynamics.

\subsection{Policy Evaluation and Improvement with Video Models}

A growing body of work explores video generation models as embodied world models, leveraging large-scale pretraining for rich visual priors~\cite{wan2025wan, agarwal2025cosmos} and pixel-space future prediction as a versatile interface across tasks, embodiments, and downstream objectives~\cite{yang2023unisimneuralclosedloopsensor, deepmind2025genie3, mei2026video, chen2025largevideoplanner}.
This makes video models attractive for policy evaluation and improvement via world model rollouts, potentially reducing reliance on expensive real-world interaction. Prior work has primarily leveraged action-conditioned video prediction in two ways: (i) generating synthetic trajectories to augment training or fine-tuning \cite{fu2026learningvideogenerationrobotic, guo2025ctrl, jang2025dreamgen, guo2026vlawiterativecoimprovementvisionlanguageaction}, and (ii) using imagined rollouts to score or optimize policies without repeated hardware trials \cite{zhu2025irasimfinegrainedworldmodel, quevedo2025worldgymworldmodelenvironment, li2025worldeval,huang2025enerverse,veorobotics2025,1x_world,tseng2025scalable,guo2025ctrl}. However, the utility of these approaches is ultimately limited by their prediction fidelity under the policy-induced distribution \cite{li2025roboticworldmodelneural}. In contact-rich robotic manipulation and high-DOF control, even small modeling errors can quickly lead to hallucinations in closed-loop rollouts, thereby resulting in inconsistent policy rankings and unstable reinforcement learning updates \cite{zhang2025worldinworldworldmodelsclosedloop}. 
We address these limitations by developing an end-to-end system for scalable data collection and model training that prioritizes broad coverage of contact-rich interaction events, building a large, diverse corpus that approximates the outcome distribution induced by arbitrary policies. 
\subsection{Leveraging Play Data in Robotics}
Play data, which consists of unstructured, task-agnostic interactions with broad coverage of interaction modes beyond narrow expert demonstrations, has been explored in prior works to enable planning \cite{lynch2019learninglatentplansplay, rosetebeas2022latentplanstaskagnosticoffline, shah2025mimicdroidincontextlearninghumanoid}, representation learning \cite{hangl2016robotic,hangl2017skilllearningautonomousrobotic, guzey2023dexterity}, policy learning \cite{cui2022playpolicyconditionalbehavior,agrawal2016learning,ebert2018visual,dasari2019robonet,levine2018learning,pinto2016supersizing}, and finetuning pretrained models~\cite{bousmalis2023robocat, walke2023don}. Recent work has also leveraged play data at a smaller scale to train world models that support policy improvement \cite{Chandra2025DiWADPA}.
These approaches typically require additional human-collected play demonstrations or policy rollouts that depend on manual supervision, which fundamentally limits scalability and the breadth of interaction diversity \cite{lynch2019learninglatentplansplay, cui2022playpolicyconditionalbehavior}. 
In contrast, our work uses fully autonomous play data collection with generalist robot policies, enabling continuous, large-scale acquisition of diverse contact-rich interactions without supervision, transforming play data into a scalable resource for training high-fidelity world models.
\begin{figure*}[t]
    \centering
    \includegraphics[width=1.0\linewidth]{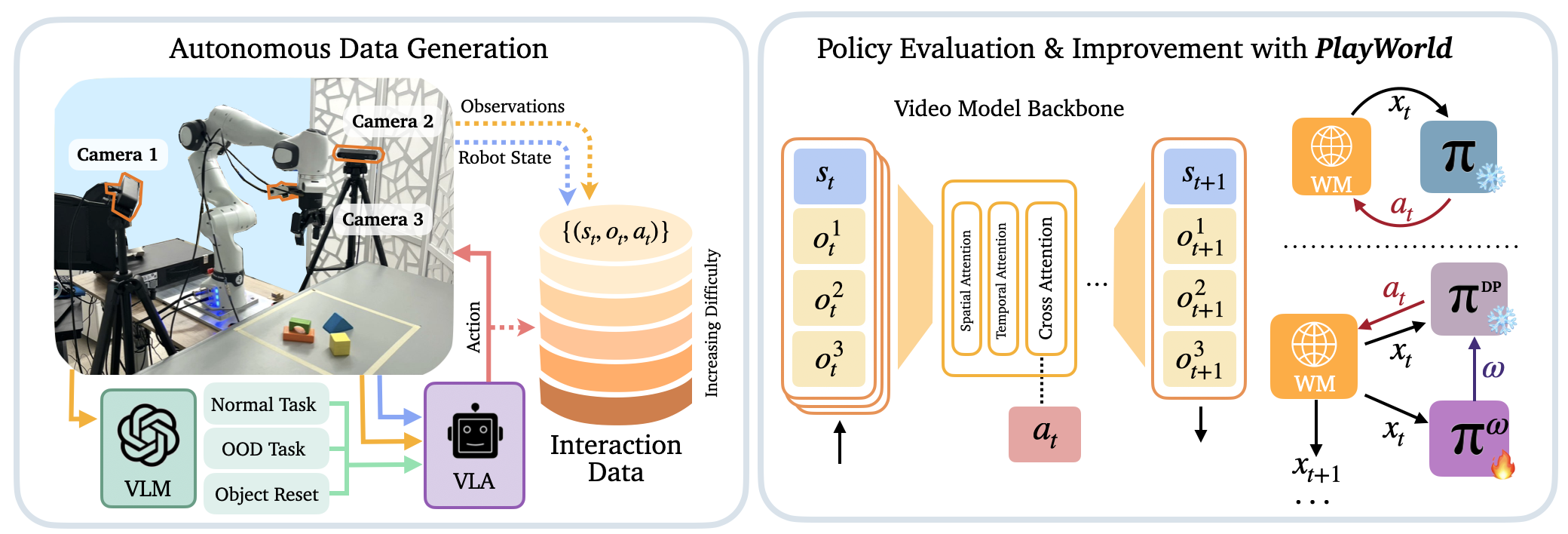}
    \vspace{-1em}
    \caption{\textbf{\algname System Diagram.} Left: Autonomous data-collection pipeline in which the VLM and VLA iteratively propose and execute tasks. Right: Video world-model backbone and the setup for policy evaluation and fine-tuning.}
    \vspace{-1em}
    \label{fig:system}
\end{figure*}
Most related to our work, SOAR~\cite{zhou2024autonomous} uses VLM-proposed tasks to autonomously collect data for improving language-conditioned policies
In contrast, we use the collected play to learn a world model, converting finite real experience into effectively unlimited imagined interactions that can more flexibly support downstream policy applications.

\section{\algname: Training Video World Models with Robot Play}
\label{sec:method}

We now introduce \emph{\algname}, a framework for training action-conditioned video models that can predict diverse contact dynamics with fine-grained precision.
In this section, we first formalize the world modeling problem as learning generalizable dynamics under distributional mismatch. 
Then, we introduce the design of our autonomous data collection system that can reliably collect hours of interaction data with no human supervision. 
Finally, we discuss model architecture and training designs that allow us to effectively learn from large quantities of heterogeneous interaction data.

\subsection{Generalizable Dynamics Learning through Play}
\label{sec:problem_formulation}

Let ${s_t \in \mathcal{S}}$ denote the proprioceptive state of the robot, ${a_t \in \mathcal{A}}$ denote actions, and ${o_t = (o_t^1, \dots, o_t^K) \in \mathcal{O}}$ denote image observations corresponding to $K$ different views (e.g., multiple overhead cameras and a wrist camera) at time $t$. We model the robot-environment dynamics using a stochastic action-conditioned video model:
\begin{equation}
    p_\theta(x_{t+1} \mid x_{1:t}, a_{1:t}),
\end{equation}
where $x_t = (s_t, o_t)$.  
As highlighted in Sec.~\ref{sec:related_work}, prior work trains such models on human demonstration data, thereby optimizing predictive accuracy only under the visitation distribution $(x_{1:t}, a_{1:t}, x_{t+1}) \sim \mathbb{D}_\text{exp}$ induced by the expert. 
As a result, the learned dynamics can suffer from distributional bias when queried under new policies and collapse to predictions that are likely under $\mathbb{D}_\text{exp}$ \cite{lu2020sample, carlini2023extracting}. 
In contrast, our goal is to train a video model that remains accurate across a broad range of policies. 

In order to achieve this, we expand the visitation distribution of the training data through data collected under a language-conditioned policy
$\pi_\text{play}(a | x, \ell)$, where $\ell \in \mathcal{L}$ denotes a natural language instruction.
By generating highly diverse language instructions in a broad range of initial conditions encountered during autonomous roll-outs, this policy yields a set:
\begin{equation}
    \mathcal{D}_{\text{play}} = \{(x_t, a_t, x_{t+1})\}_{a_t \sim \pi(\cdot \mid x_t, \ell), \, \ell \sim p(\mathcal{L})}
\end{equation}
that can better approximate visitation distributions induced by a wide range of policies.

\subsection{Autonomous Robot Play Data Collection}
\label{sec:play data collection}

Next, we discuss a practical instantiation of $\pi_\text{play}$ that can enable reliable and continuous autonomous data collection on hardware. 
In order to obtain useful and scalable play data, we outline three key design requirements:
\begin{enumerate}
    \item The robot needs to engage in \emph{diverse interaction} with objects in the scene (in order to obtain meaningful coverage as data collection expands)
    \item The system must \emph{reliably} prevent and recover from potential failures when human supervision is unavailable
    \item The system should be amenable to diverse objects and \emph{generalize} to diverse language instructions $\ell$ without requiring manual engineering
\end{enumerate}

Building on these requirements, we develop a practical system atop the DROID manipulation setup~\cite{khazatsky2025droidlargescaleinthewildrobot} that requires no additional hardware modifications and supports up to eight hours of continuous, fully autonomous data collection.
As illustrated in Fig.~\ref{fig:system}, our system is composed of two interacting components: a \emph{task proposer} that generates tasks to engage the robot in meaningful interactions with objects, and a \emph{task executer} that performs the task to generate interaction data.

\smallskip
\noindent \textbf{Task Proposer.} 
We instantiate the proposer using a vision–language model (VLM)~\cite{openai2024gpt4technicalreport} that generates diverse, scene-grounded instructions from the robot’s current image observation(s) $o_t$. 
At the start of each episode, the VLM is prompted to produce a natural-language instruction $\ell$ (e.g., ``push the object forward'' or ``stack the object'').
To promote data diversity while preserving executability, we prompt the VLM to add small perturbations to nominal commands (e.g., alternative verbs or object descriptions) to probe different behavior modes.
Compared to novelty- or uncertainty-driven on-policy exploration, our language-conditioned play requires no reward design or auxiliary training, and generates semantically grounded diversity by varying task intent (rather than injecting action noise), yielding visitation distributions that better match test-time policy behavior.
More details on implementation and example prompts can be found in Appendix~\ref{appendix:implementation}.

\smallskip
\noindent \textbf{Task Executer.} 
We use a vision–language–action (VLA) policy \cite{intelligence2025pi05visionlanguageactionmodelopenworld} as the executer that can be conditioned on arbitrary natural-language instructions. While it succeeds on nominal commands (e.g., “pick up the object”), small perturbations in the instruction such as verb choice or unfamiliar object descriptions often lead to large variances in policy behavior \cite{fei2025libero}. We exploit this as a source of diversity: during long-horizon autonomous roll-outs, we intentionally randomize instructions under naturally-induced varied initial object poses, boosting the diversity of contact dynamics modes for interaction-rich data collection.

\smallskip
\noindent \textbf{Safety Filter and Resets.} 
To enable reliable unsupervised execution for long periods of time, we utilize a lightweight safety filter that constrains the robot to operate within conservative workspace limits. 
Specifically, we enforce per-joint limits to prevent drastic motions, and prompt the VLM to detect when any object is drifting towards the boundary of the robot's reachability limit, to instruct the VLA to ``reset" the scene by retrieving that object back to the reachable workspace. 
Since the setup is task/object-agnostic, during play data collection involving large number of objects (such as in Sec.~\ref{exp:scaling}), we can simply add or remove arbitrary objects during collection. This simple design works well in practice and allows us to collect data for long durations including overnight.

\subsection{Model Architecture and Training} 
\label{sec:training}
Finally, we discuss our choice of world model architecture and a curriculum-based training scheme for learning from uncurated interaction data. 

\smallskip
\noindent\textbf{Video Model Architecture.}
Given the dataset $\mathcal{D}_\text{play}$, our goal is to train a world model that produces high-fidelity visual outputs that simulate fine-grained interactions with the environment.
Following \cite{guo2025ctrl}, we adopt a pre-trained stable video diffusion (SVD) backbone \cite{blattmann2023stable} with factorized spatial and temporal attention, allowing us to inject fine-grained per-frame action conditioning to disentangle motion signals from frame appearances for strong controllability.
We train our model to jointly predict three camera views to minimize the influence of partial observability, similar to \cite{veorobotics2025, kim2026cosmos}, and initialize our model with weights pre-trained on the DROID dataset \cite{khazatsky2025droidlargescaleinthewildrobot}.
Then, we fine-tune the world model on the play dataset $\mathcal{D}_{\text{play}}$ using the diffusion loss function:
\begin{equation}
    \label{eq:diffusion_loss_fn}
    L_{\boldsymbol{\epsilon}} = \mathbb{E}_{\mathbf{x}_0, t, \boldsymbol{\epsilon}} 
    \left[
    \left\|
    \boldsymbol{\epsilon} - \boldsymbol{\epsilon}_\theta(\mathbf{x}_t, t)
    \right\|^2
    \right],
\end{equation}
where ${\boldsymbol{\epsilon} \sim \mathcal{N}(0, \boldsymbol{I})}$.
We train the full \algname model on 8x H200 GPUs with batch size of 64 for two days.


\smallskip
\noindent\textbf{Curriculum Learning.} As we scale up play data collection to increase coverage of diverse states and transitions, the resulting dataset introduces two challenges: (1) high redundancy, where the majority of data are dominated by very similar transitions, and (2) multi-modality, where there also exists many rarely-occurring transitions that exhibits strong diversity (long-tail) \cite{cui2022playpolicyconditionalbehavior}. 
This unbalanced and long-tailed distribution makes learning with standard training procedures difficult, causing models to easily overfit to simple patterns while failing to capture rare interactions \cite{hu2020learningsegmenttail, xing2025shortcutlearninggeneralistrobot}.

To enable a more balanced exposure, we adopt a curriculum learning setup \cite{bengio2009curriculum, wen2025light} to feed training data into the model in order of (auto-rated) ``difficulty": initializing with frequently occurring free space motions and static contacts, and gradually biasing sampling towards rare, harder-to-learn interactions. 
Concretely, suppose we are given a small set of human-collected demonstrations $\mathcal{D}_{\text{exp}} = \{ \tau_i \}_{i=1}^{N_s}$, where each trajectory $\tau = \{(x_t, a_t)\}_{t=1}^T$ corresponds to successful task execution. Let $\mathcal{D}_{\text{play}} = \{ \tilde{\tau}_j \}_{j=1}^{N_p}$ denote uncurated and unlabeled play data collected using the protocol from Sec.~\ref{sec:play data collection}. Let $f_{\text{clip}}(\cdot)$ denote a frozen CLIP~\cite{radford2021learning} image encoder.
For each observation $o_t$, we compute an embedding
\begin{equation}
    z_t = f_{\text{clip}}(o_t) \in \mathbb{R}^d . 
\end{equation}

We first extract a set of representative \emph{success centroids} from $\mathcal{D}_{\text{exp}}$ using all embeddings from success trajectories,
$\mathcal{Z}_{\text{succ}} = \{ z_t \mid o_t \in \mathcal{D}_{\text{succ}} \}$,
and then apply $K$-means clustering to obtain centroids
$\mathcal{C} = \{ c_k \}_{k=1}^K , \quad c_k \in \mathbb{R}^d$.
These centroids serve as prototypes of task-relevant transitions observed in successful executions.
For each observation $o$ in $\mathcal{D}_\text{play}$, we define its \emph{distance-to-success} score as
\begin{equation}
    d(o) = \min_{c_k \in \mathcal{C}} \, \| f_{\text{clip}}(o) - c_k \|_2 .
\end{equation}
Intuitively, smaller values of $d(o)$ indicate transitions that are visually and semantically closer to successful trajectories, while larger values correspond to less task-relevant or exploratory interactions.
We then induce a curriculum by partitioning $\mathcal{D}_{\text{play}}$ into $R$ disjoint ranks based on this distance:
\begin{equation}
    \mathcal{D}_{\text{play}} = \bigcup_{r=1}^R \mathcal{D}^{(r)}, 
    \quad \mathcal{D}^{(r)} = \{ o \mid d(o) \in [\delta_{r-1}, \delta_r) \},
\end{equation}
where $\{ \delta_r \}_{r=0}^R$ are distance thresholds chosen such that
$\delta_0 = 0$ and $\delta_R = +\infty$.
Lower ranks correspond to samples that are closer to success trajectories,
while higher ranks contain increasingly out-of-distribution interactions.
During training, we progressively expand the sampling distribution over ranks, starting from lower $r$ and gradually incorporating higher-ranked samples.
This curriculum mitigates redundancy in free-space motion and counteracts the long-tailed distribution of critical interaction dynamics, enabling the model to learn rare but essential transitions more effectively. 

\section{Experiments}
\label{sec:evaluation}

\begin{figure*}[t]
    \centering

    \captionsetup{skip=2pt}

    \begin{minipage}{\textwidth}
        \centering
        \captionof{table}{Per-category perceptual similarity metrics on interaction-centric benchmark. \algname improves prediction quality on contact-rich failure modes, with further gains from scaling and curriculum learning.}
        \label{tab:main_results}
        \resizebox{\textwidth}{!}{
        \begin{tabular}{lcccccccccccc}
            \toprule
            \textbf{Training Mix}
            & \multicolumn{2}{c}{Success}
            & \multicolumn{2}{c}{Missed Grasp}
            & \multicolumn{2}{c}{Slide}
            & \multicolumn{2}{c}{Slip}
            & \multicolumn{2}{c}{Deformation}
            & \multicolumn{2}{c}{Collision} \\
            \cmidrule(lr){2-3}
            \cmidrule(lr){4-5}
            \cmidrule(lr){6-7}
            \cmidrule(lr){8-9}
            \cmidrule(lr){10-11}
            \cmidrule(lr){12-13}

            & LPIPS & SSIM
            & LPIPS & SSIM
            & LPIPS & SSIM
            & LPIPS & SSIM
            & LPIPS & SSIM
            & LPIPS & SSIM \\
            \midrule
            Human Demo (6h)
                & 0.084 & 0.867 & 0.080 & 0.875 & 0.090 & 0.850 & 0.090 & 0.865 & 0.108 & 0.820 & 0.086 & 0.852 \\
            Human Play (6h)
                & 0.086 & 0.861 & 0.071 & 0.869 & 0.089 & 0.864 & 0.088 & 0.867 & 0.100 & 0.831 & 0.080 & 0.873 \\
            Robot Play (6h)
                & 0.082 & 0.870 & 0.066 & 0.883 & 0.077 & 0.865 & 0.078 & 0.871 & 0.099 & 0.831 & 0.074 & 0.888 \\
            \cmidrule(lr){1-13}
            Robot Play (30h)
                & 0.071 & 0.873 & \cellcolor{mygreen!25}0.064 & 0.887 & 0.073 & 0.876 & 0.072 & 0.879 & 0.094 & 0.833 & 0.076 & 0.883 \\
            Robot Play (Curriculum)
                & \cellcolor{mygreen!25}0.070 & \cellcolor{mygreen!25}0.880 & 0.066 & \cellcolor{mygreen!25}0.890 & \cellcolor{mygreen!25}0.071 & \cellcolor{mygreen!25}0.890 & \cellcolor{mygreen!25}0.070 & \cellcolor{mygreen!25}0.884 & \cellcolor{mygreen!25}0.093 & \cellcolor{mygreen!25}0.836 & \cellcolor{mygreen!25}0.072 & \cellcolor{mygreen!25}0.893 \\
            \bottomrule
        \end{tabular}
        }
    \end{minipage}
    \vspace{1.2em}
    \begin{minipage}{\textwidth}
        \centering
        \vspace{0.6em}
        \includegraphics[width=\linewidth]{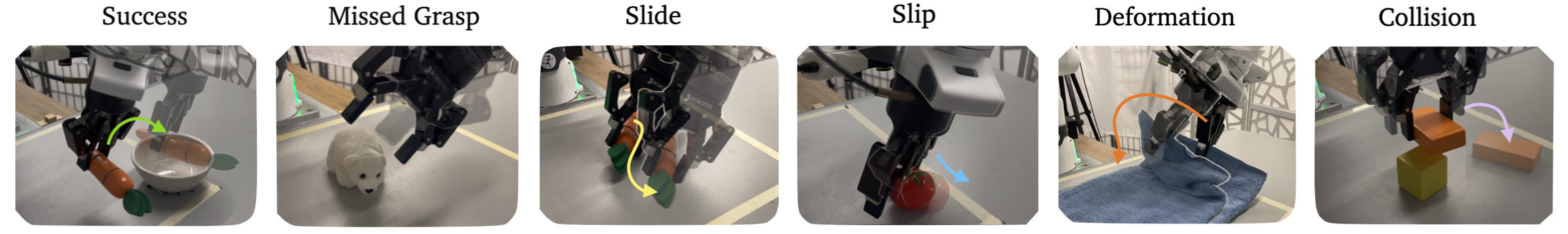}
        \captionof{figure}{Illustration of each test category from the interaction-centric benchmark in Table~\ref{tab:main_results}.}
        \label{fig:category}
    \end{minipage}
    \vspace{-1.0em}
\end{figure*}

In this section, we design experiments to explore the following questions:

\begin{enumerate}
\item Does \algname induce more diverse object interactions compared to human-collected data (Sec.~\ref{exp:coverage})?
    \item Can \algname improve video prediction accuracy for diverse object interactions compared to models trained on human demonstration data (Sec.~\ref{exp:dynamics-prediction})?
    \item Can \algname enable fine-grained policy evaluation by reliably predicting outcomes across a broad range of policies and tasks (Sec.~\ref{exp:policy-eval})?
    \item Can \algname enable policy fine-tuning through interactive roll-outs in the video model (Sec.~\ref{exp:finetune})?
    \item Does \algname result in improved accuracy and generalization with data scale compared to human-collected data (Sec.~\ref{exp:scaling})?
\end{enumerate}

\subsection{Experiment Setup}
\label{exp:setup}

\smallskip
\noindent\textbf{Evaluation Setup.} 
To evaluate whether \algname can produce more accurate dynamics predictions under realistic settings where object interactions are diverse and contact-rich, we carry out experiments on three distinct object sets featuring objects with different physical properties and interaction behaviors, each with a collection of possible tasks:

\begin{itemize}
    \item Set 1: Bowl, carrot, polar bear (put the carrot/polar bear into/out of the bowl)
    \item Set 2: Rectangular block and cube (stack/unstack the block on top of the cube)
    \item Set 3: Towel (fold/unfold towel)
\end{itemize}

For our method (\algname), we follow Sec.~\ref{sec:play data collection} to collect a total of 30 hours of task-agnostic autonomous robot play data containing diverse robot-object interactions combined across all 3 object sets. 

\smallskip
\noindent\textbf{Baselines.} We collect two types of teleoperation data as baselines: \textit{human demo} data consisting of task-specific expert demonstrations, and \textit{human play} data where the operator is instructed to freely interact with given objects in a task-agnostic manner. In total, we collect 6 hours of  human demo data on the identical setup as robot play data.
We initialize all models from the DROID-pretrained checkpoint in \cite{guo2025ctrl}. 
Our \textit{human~demo} baseline corresponds to the standard demonstration-only fine-tuning paradigm (as in \cite{guo2025ctrl, quevedo2025worldgymworldmodelenvironment}) designed in particular for our environment, enabling a controlled comparison that isolates the effect of interaction data.

\subsection{Does PlayWorld Induce Diverse Interactions?}
\label{exp:coverage}
\begin{wrapfigure}[19]{r}{0.5\textwidth}
    \centering
    \vspace{-1.2\baselineskip}
    \includegraphics[width=\linewidth]{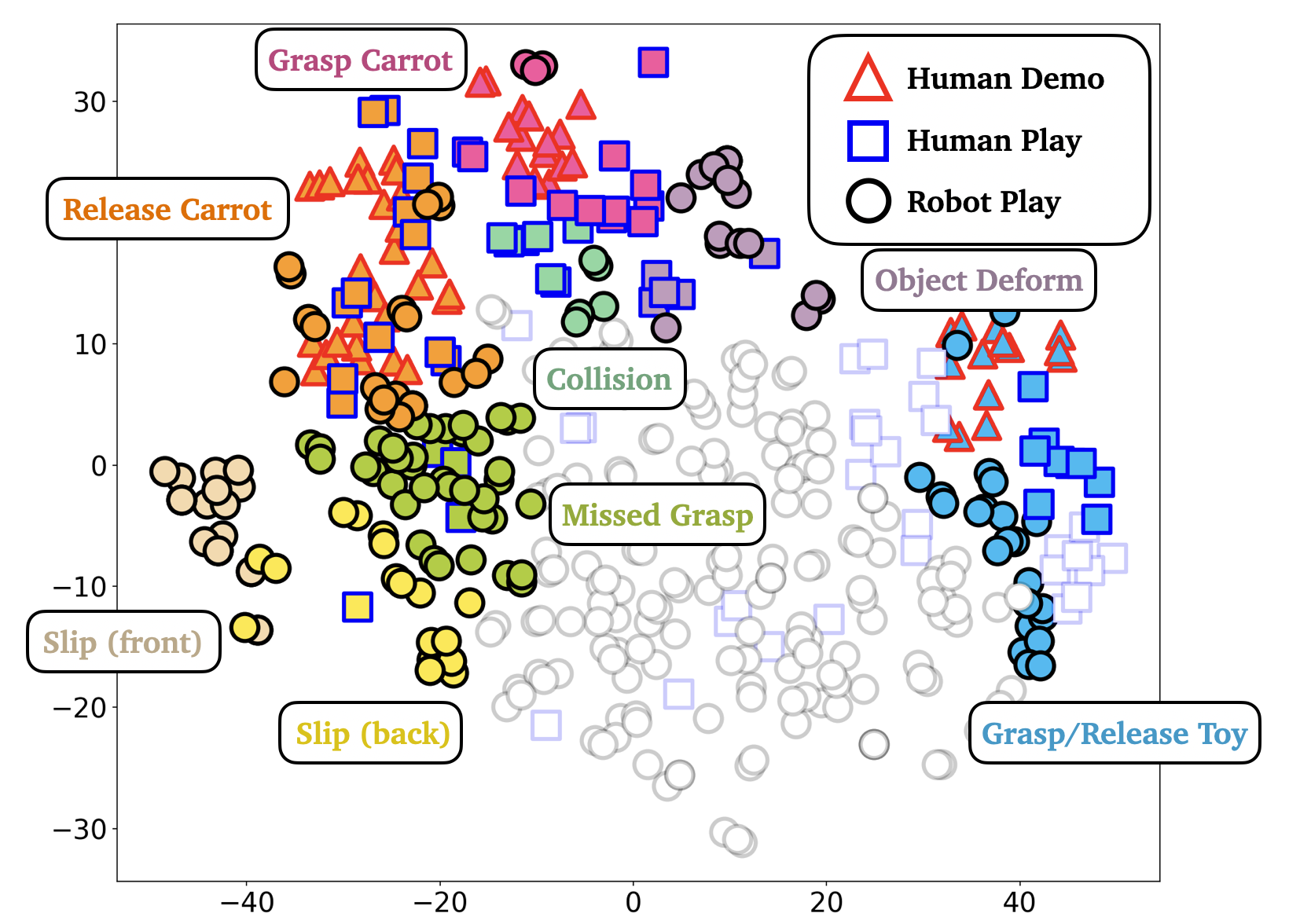}
    \caption{\textbf{t-SNE analysis of training samples}. Robot play data exhibits markedly broader behavioral coverage than human-collected trajectories. Colors indicate coarse interaction modes assigned by a human annotator. }
    \label{fig:tsne}
    \vspace{-4em}
\end{wrapfigure}

First, we evaluate if \algname can induce more diverse interactions with objects compared to human-collected data, as outlined in Sec.~\ref{sec:problem_formulation}. For this comparison, we analyze a subset of autonomous play data alongside the \emph{human~demo} and \emph{human~play} baselines from object set 1 (Sec.~\ref{exp:setup}) and provide a t-SNE visualization~\cite{van2008visualizing} on CLIP embeddings of the image observations. 
To better interpret the embedding space, we let a human annotator inspect the corresponding video snippet for each data point and assigns it to a coarse, human-interpretable interaction mode when applicable.
As shown in Fig.~\ref{fig:tsne}, autonomous play exhibits substantially broader behavioral coverage, capturing diverse contact-rich events that are consistent with human annotations (e.g., missed grasps, collisions, and slips). In contrast, human-collected data forms a more concentrated cluster that largely reflects successful transitions.
 More results on conducting data clustering at larger scales can be found in Appendix~\ref{appendix:play_data}.
 
\subsection{Can \algname Accurately Predict Contact Dynamics?}
\label{exp:dynamics-prediction}

\begin{figure*}
    \centering
    \includegraphics[width=0.95\linewidth]{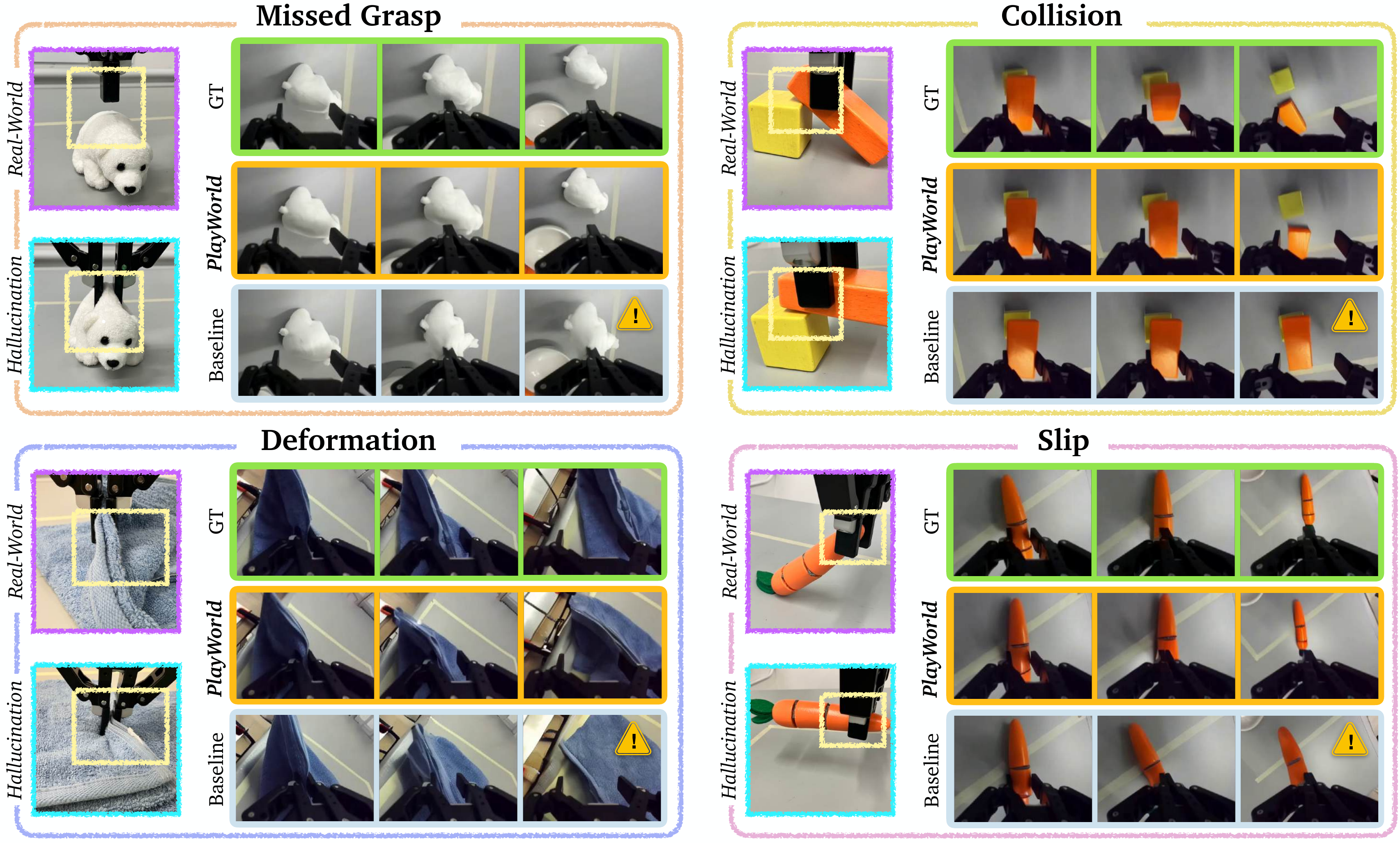}
    \caption{\textbf{PlayWorld faithfully captures fine-grained dynamic behaviors.} \algname's predictions closely match fine-grained dynamic interaction modes in the real-world, such as missed grasps, collisions, deformations, and slips.
    }
    \label{fig:result:replay}
\end{figure*}

Next, we evaluate if the \algname's broader data coverage enables training better video world models for predicting challenging object dynamics.
To this end, we construct an interaction-centric benchmark with $500+$ clips sampled from roll-outs generated by a diverse collection of 20+ robot policies, categorized into 6 behavior modes by human annotators (Fig.~\ref{fig:category}), containing both successful task executions and common failure modes that are critical in determining policy behavior.
Given a short ground-truth observation–state trajectory as conditioning input, we roll out the ground-truth actions in the world model and compare the generated vs. ground-truth future frames using visual metrics such as LPIPS~\citep{zhang2018unreasonableeffectivenessdeepfeatures} and SSIM~\citep{wang2004image} (results on other visual metrics can be found in Appendix~\ref{appendix:additional_results:replay}). 

As shown in Table~\ref{tab:main_results}, while the prediction quality for successful interactions is similar across different training mixtures, \algname provides consistent improvements for other dynamic interactions. In addition, scaling PlayWorld data from 6h to 30h further boosts performance. Notably, curriculum learning offered substantial advantage in improving prediction quality for more dynamic interactions, given the same training data.
Fig.~\ref{fig:result:replay} provides some example predictions when rolling out a given action sequence from the ground-truth trajectory under different world models; we observe that predictions from the baseline methods often collapse to behaviors biased by the training data, with ``hallucinated success" being the most common failure mode.

\subsection{Can \algname Predict the Performance of Different Policies?}
\label{exp:policy-eval}

\begin{wrapfigure}[14]{r}{0.44\textwidth}
    \centering
    \vspace{-1.4\baselineskip}
    \includegraphics[width=\linewidth]{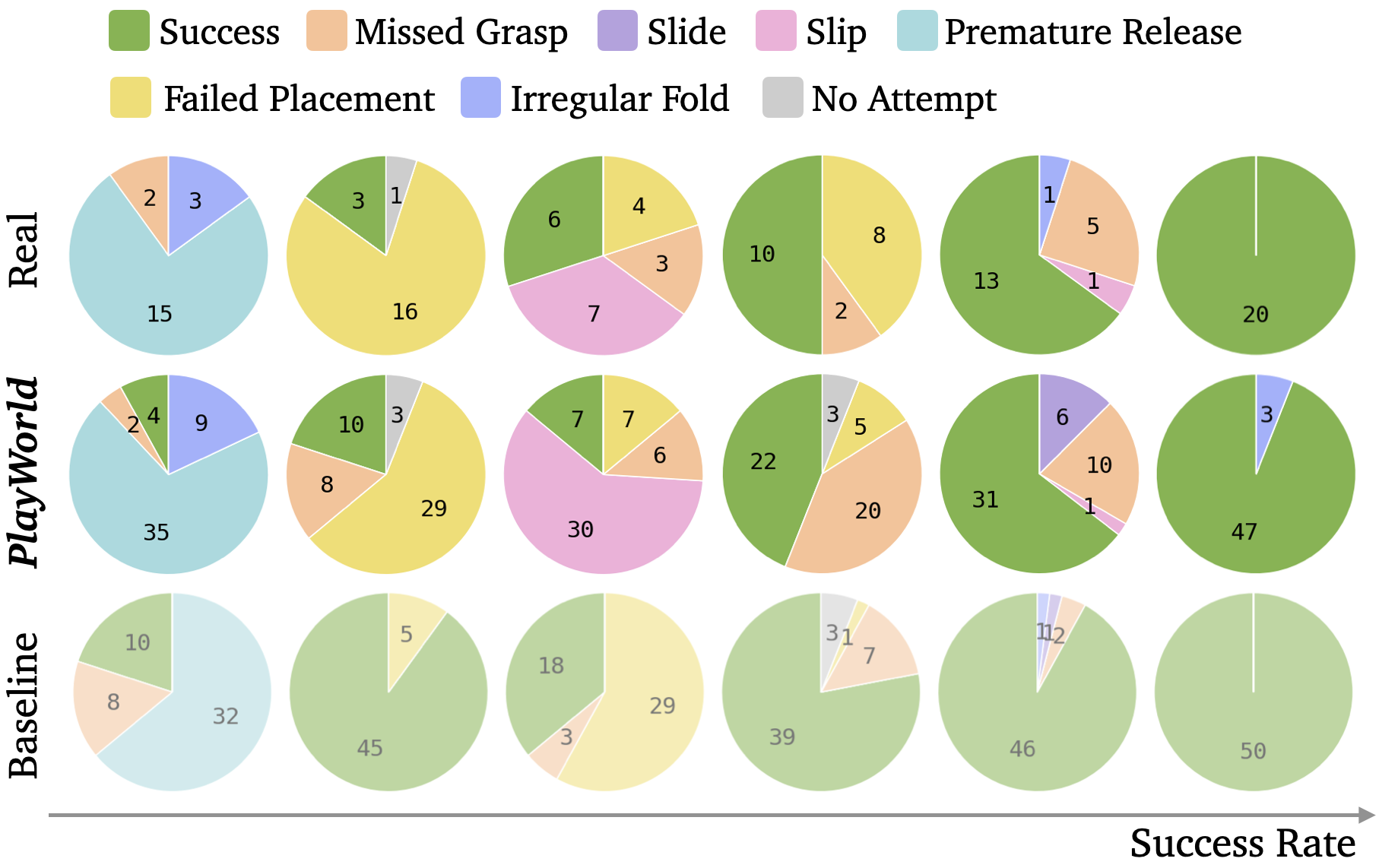}
     \caption{\algname’s predicted policy outcome distribution matches closely with real-world behavior. 
     }
    \label{fig:failure_analysis}
    \vspace{-10em}
\end{wrapfigure}

While high-level policy capabilities such as instruction following are relatively easy to assess, evaluating fine-grained visuomotor abilities is substantially harder: small errors in contact timing, object position, or contact forces can lead to qualitatively different outcomes during closed-loop execution. 
To rigorously test whether \algname can faithfully predict policy performance, we construct a diverse suite of 18 policies by (i) training diffusion policies (DPs) \cite{chi2025diffusion} from scratch and (ii) fine-tuning $\pi_0$~\cite{black2024pi_0} using human demonstrations that vary in both quantity and quality. This produces policies spanning a wide range of success rates and exhibiting diverse failure modes (Fig.~\ref{fig:failure_analysis}), providing a sensitive probe of whether a world model captures subtle contact-rich dynamics under realistic settings. 
For each policy, we perform 20 real-world experiments and 50 simulated experiments in each world model.

\begin{figure*}[t]
    \centering
    \includegraphics[width=\linewidth]{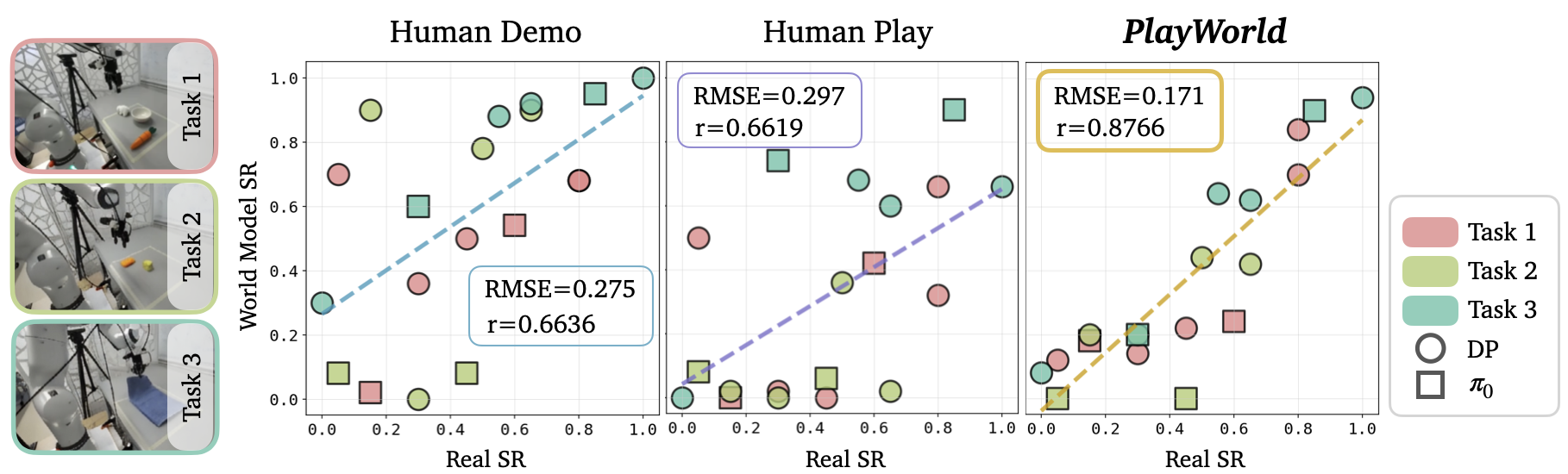}
    \caption{\textbf{Policy Evaluation Success Rate (SR) Correlation.} Across diverse policies (architectures, training mixtures, and tasks), \algname’s predicted success rates are most strongly correlated with observed success rates and show the lowest variability.
    }
    \label{fig:result:eval_correlation}
\end{figure*}

\begin{figure*}[t]
    \centering
    \includegraphics[width=\linewidth]{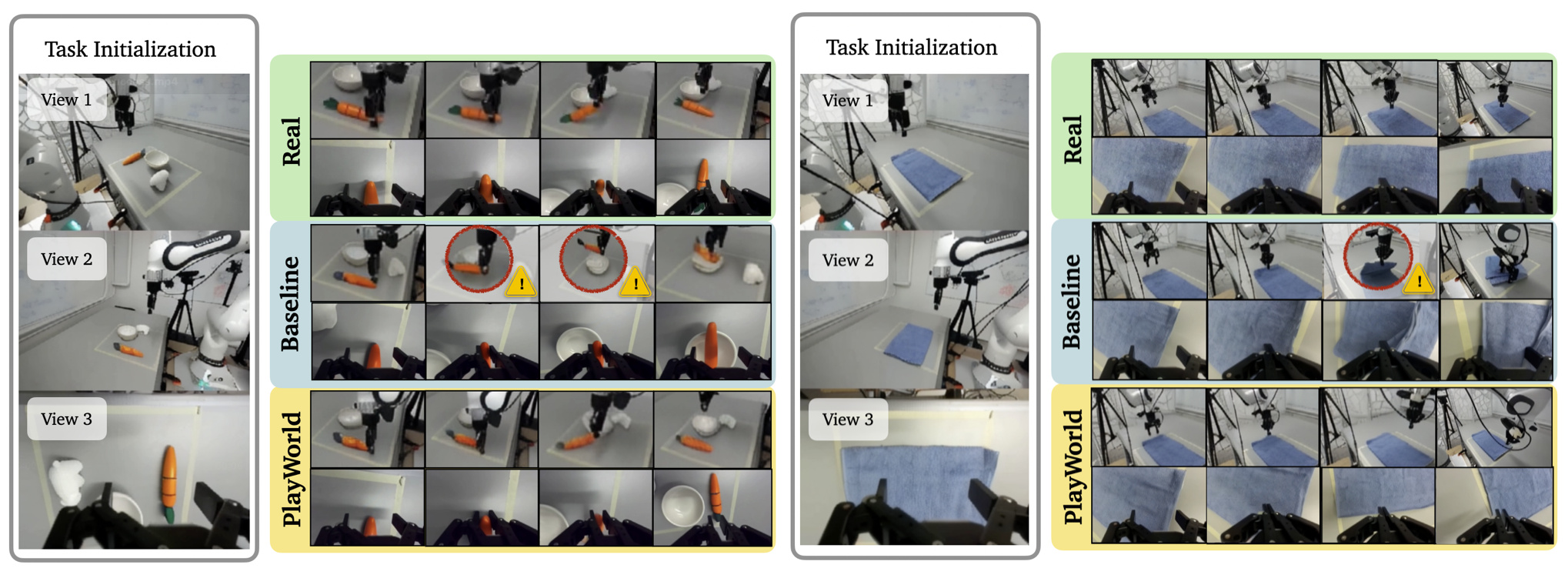}
    \caption{\textbf{Example Policy Interaction with different World Models.} We provide closed-loop rollouts with two example policies to demonstrate how hallucinated success can impact prediction outcomes.}
    \label{fig:result:eval_rollout}
\end{figure*}

As shown in Fig.~\ref{fig:result:eval_correlation}, \algname generalizes well to a broad family of policies with substantially different success rates and behaviors (despite training on roll-outs collected form a single autonomous play policy). Predicted success rates from PlayWorld are strongly correlated with real-world success rates (Pearson correlation: 0.8766). 
In contrast, baseline models are only capable of capturing a narrow set of failure modes.
When policies induce interaction patterns that are underrepresented in their training distributions, baseline models tend to either regress toward familiar outcomes or generate unrealistic dynamics with noticeably degraded visual quality (Fig.~\ref{fig:result:eval_rollout}), leading to large discrepancies between predicted and observed success rates.
In particular, we find that human play data yields the worst performance and often produces blurry predictions, which suggests that out-of-distribution diversity that doesn't match well with policy behavior could hurt performance.

Beyond aggregate success rates, we further assess whether \algname can produce meaningful predictions of both success and failure by comparing human-annotated failure modes for each set of experiments. In Fig.~\ref{fig:failure_analysis}, we present real and predicted failure-mode distributions for several representative policies and find that \algname's predicted distributions broadly align with empirical outcomes, whereas the behavior within the baseline models have large variances. 

\subsection{Policy Fine-Tuning in the World Model}
\label{exp:finetune}

Many reinforcement learning fine-tuning algorithms can substantially improve policy robustness, but are seldom deployed on physical robots because real-world interaction is expensive, safety-limited, and slow to run.
While many existing video world models are used to optimize policies by generating synthetic data for supervised fine-tuning \cite{jang2025dreamgen, wang2025learningrealworldactionvideodynamics, xu2025vilpimitationlearninglatent}, here we show that with  high dynamics prediction quality, \algname can effectively support in-model RL fine-tuning that confers substantial gains in real-world deployment with minimal sim-to-real gap, making real-world RL more practical and scalable.

\begin{wrapfigure}{r}{0.58\textwidth}
    \includegraphics[width=\linewidth]{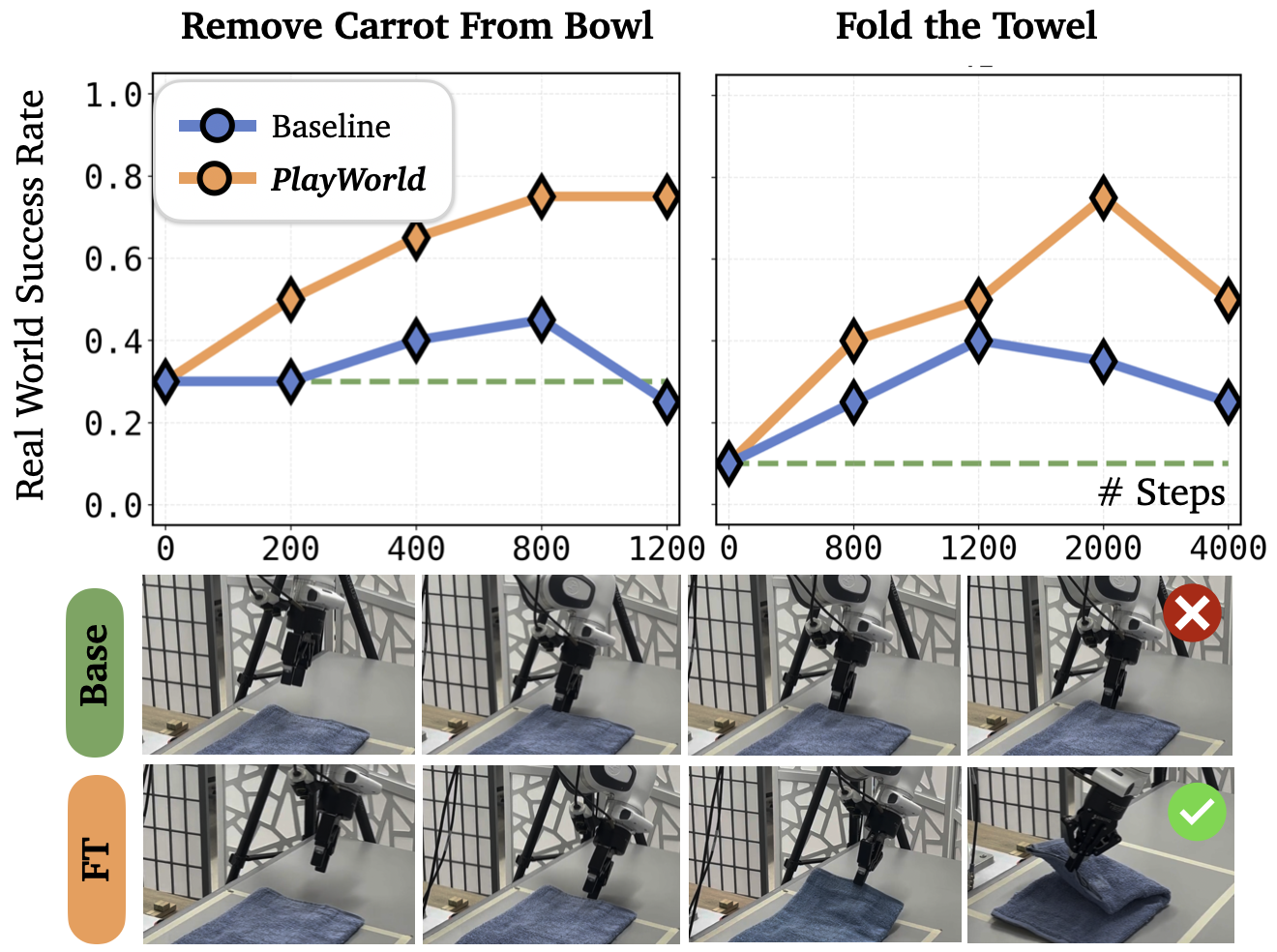}
    \caption{\textbf{Fine-tuning Experiment Results.} \algname-finetuned policy demonstrates improved success rates and learns robust recovery behaviors beyond seen demonstrations.}
    \label{fig:finetune}
\end{wrapfigure}
To support stable fine-tuning, we adopt Diffusion Steering via Reinforcement Learning (DSRL) from \cite{wagenmaker2025steering}, which freezes the base diffusion policy and instead learns a lightweight latent-noise policy $\pi^w$ that samples the initial diffusion noise $w$ (replacing $w \sim \mathcal{N}(0,I)$) to steer the resulting action $a=\pi^{W}_{dp}(s,w)$ toward higher reward, while avoiding unstable backpropagation through the multi-step denoising chain.
We train a simple progress-based reward function on small amounts of demonstration data, and compute the difference in predicted progress at each frame as dense rewards. 
Despite the simplistic setup, we observe surprisingly effective fine-tuning results.
Fig.~\ref{fig:finetune} summarizes our results on performing fine-tuning over two example tasks. Starting from a base diffusion policy trained with fewer than 10 demonstrations,
performing online fine-tuning fully within \algname improves real-world success rates by up to 65\%, compared to fine-tuning in the baseline video model.
In particular, we find that the fine-tuned model performs much more robustly on out-of-distribution initializations (including ones unseen during fine-tuning) and learns subtle recovery behaviors like scooping, as illustrated at the bottom of Fig.~\ref{fig:finetune}.
In contrast, the baseline model trained on demonstrations offers less significant improvements and becomes unstable as the policy tries to ``hack" the world model when failure modes are incorrectly identified, resulting in decrease in real-world success rates.
We present more detailed findings in the Appendix~\ref{app:additional_finetuning}.

\subsection{Scaling and Generalization}
\label{exp:scaling}

Finally, we evaluate whether PlayWorld delivers sustainable, scalable data collection for world-model learning by measuring continued performance gains as we increase \emph{data scale} and \emph{object diversity}.

\begin{wrapfigure}{r}{0.44\textwidth}
    \centering
    \vspace{-0.5\baselineskip}
    \includegraphics[width=\linewidth]{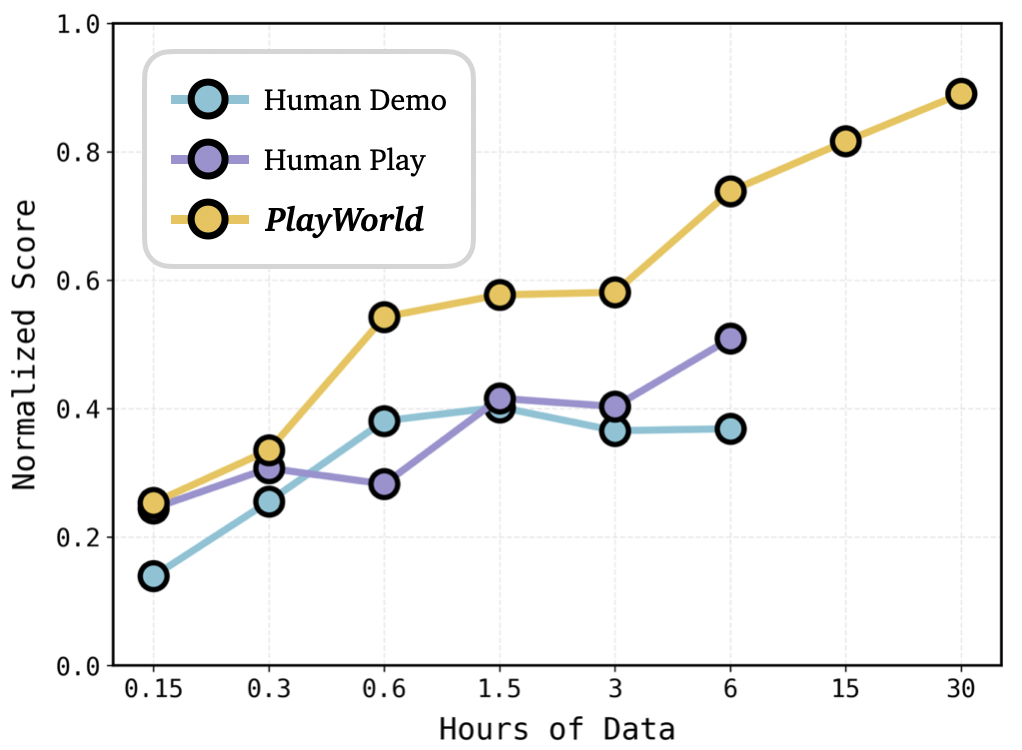}
    \caption{Scaling behavior for world models trained with different data mixtures.}
    \label{fig:scaling}
    \vspace{-3em}
\end{wrapfigure}
\smallskip
\noindent\textbf{Data Scaling.} While scaling up data might appear to be a straightforward recipe for training better models, we show that the effectiveness of scaling depends strongly on data coverage. As shown in Fig.~\ref{fig:scaling}, the video model trained with \algname continues to improve on visual-quality metrics as we scale the dataset up to 30 hours. By contrast, baseline models show only marginal gains at comparable scale.
This divergence highlights a central advantage of autonomous play: rather than repeatedly sampling narrow, success-focused trajectories, play data expands coverage over contact events, failure modes, and counterfactual transitions. As a result, additional data meaningfully broadens the learned dynamics model instead of reinforcing existing biases. These results suggest that PlayWorld supports effective scaling, where increased data directly translates into improved generalization.

\begin{figure}
\begin{minipage}{0.48\linewidth}
    \centering
    \includegraphics[width=\linewidth]{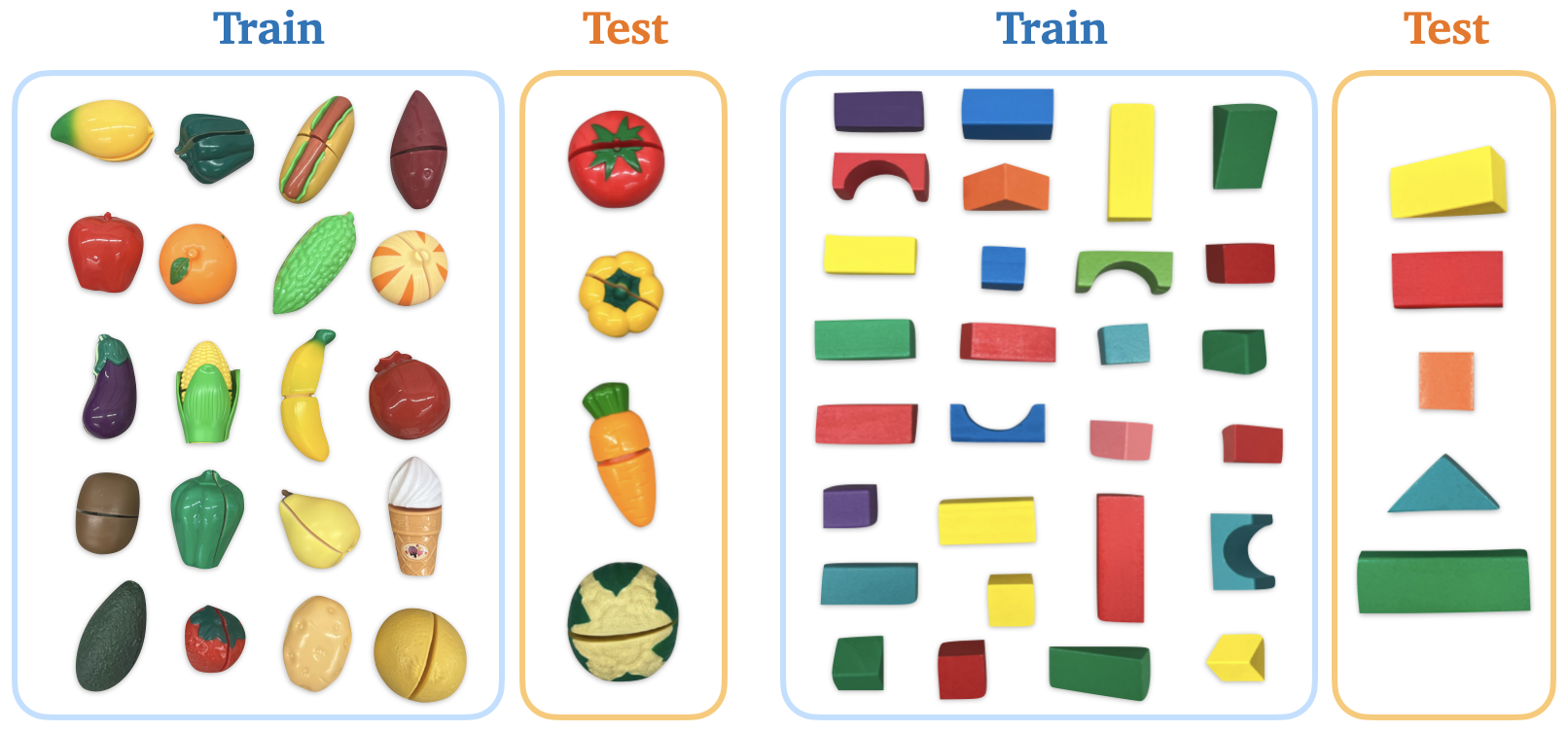}
\end{minipage}
\hfill
\begin{minipage}{0.48\linewidth}
    \centering
    \small
    \setlength{\tabcolsep}{4pt}
    \begin{tabular}{lcccc}
    \toprule
    \textbf{Objects} &
    \multicolumn{2}{c}{\textbf{Fruit}} &
    \multicolumn{2}{c}{\textbf{Blocks}} \\
    \cmidrule(lr){2-3} \cmidrule(lr){4-5}
    & LPIPS $\downarrow$ & SSIM $\uparrow$ & LPIPS $\downarrow$ & SSIM $\uparrow$ \\
    \midrule
    0\%   & 0.0991 & 0.8333 & 0.0978 & 0.8501 \\
    33\%  & 0.0794 & 0.8612 & 0.0713 & 0.8810 \\
    66\%  & 0.0761 & 0.8697 & 0.0684 & 0.8853 \\
    100\% & \cellcolor{mygreen!25}0.0705 & \cellcolor{mygreen!25}0.8757 & \cellcolor{mygreen!25}0.0640 & \cellcolor{mygreen!25}0.8948 \\
    \bottomrule
    \end{tabular}
\end{minipage}
\captionof{figure}{\textbf{Object Generalization Results.} We observe strong improvements in prediction quality of interaction with unseen objects as we include more diverse items in the training data.}
\label{fig:objects_side}
\end{figure}

\smallskip
\noindent\textbf{Object Generalization.}
We next test whether increasing object diversity in the training corpus improves transfer to unseen objects.
As shown in Fig.~\ref{fig:objects_side}, as we increase the fraction of training objects from 0\% to 100\%, prediction quality improves consistently on held-out objects across both test categories, despite substantial shifts in appearance and geometry.
This suggests that \algname learns dynamics that transfer across object instances, rather than memorizing object-specific visual features.
We attribute this generalization to PlayWorld’s ability to scale object diversity while still inducing rich contact interactions during play, providing supervision on shared interaction patterns (e.g., contact, slip, deformation) that recur across different objects.

\section{Conclusion} 
\label{sec:conclusion}

We present \algname, an efficient, autonomous, and scalable pipeline for training high-fidelity action-conditioned video world models on interaction-rich robot play data. 
By collecting large volumes of diverse experience autonomously (including unattended overnight operation), \algname yields broader coverage of contact events and a richer learning signal than success-biased demonstrations.
The resulting video world model achieves strong dynamics fidelity on both nominal and off-nominal interactions, enabling reliable fine-grained policy evaluation and in-model RL fine-tuning.
Finally, we observe continued gains with increased data scale and object diversity, suggesting a practical path toward high-quality robot simulators that support policy evaluation and improvement beyond hardware constraints.

\section{Limitations and Future Work}

\smallskip
\noindent \textbf{Improving Data Collection.}
While \algname can generate many diverse interaction data at scale, the current collection strategy does not explicitly optimize for sample efficiency and can produce redundant trajectories.
Future work could explore better active data collection strategies that prioritize high uncertainty or underrepresented interactions that can further reduce hallucinations and improve physical consistency. Additionally, the current system relies on the availability of a sufficiently capable play policy; designing specialized or adaptive play policies that explicitly target rare contact events could further enhance coverage and efficiency.

\smallskip
\noindent \textbf{Dynamic Discrepancies.}
Despite significant improvements, PlayWorld does not eliminate hallucinations entirely. Prediction errors can still arise from open-loop rollout discrepancies, fixed action horizons, and mismatches between the control modes used during data collection and policy evaluation, which can accumulate over long horizons and degrade closed-loop stability. Future work could investigate more flexible prediction horizons, hierarchical or receding-horizon rollouts, and tighter integration between action conditioning and control representations to better align imagined dynamics with real-world execution.

\smallskip
\noindent \textbf{Scaling Up.}
While we demonstrate strong scaling trends within our experimental setup, extending PlayWorld beyond a controlled lab environment remains an open challenge. Scaling to more diverse objects, scenes, robot embodiments, and real-world settings will require improved strategies for learning from increasingly heterogeneous data distributions. As data scale increases, improved curriculum design and data balancing mechanisms will become even more important; developing principled methods for automatically defining curriculum that optimize both prediction fidelity and robustness to long-tail interactions remains a key direction for future work. Efficiently leveraging large-scale, multi-robot datasets and potentially incorporating non-robot video data for broader physical priors also presents exciting opportunities.

\vspace{5pt}

Overall, we believe that PlayWorld showcases the promise of scaling autonomous play data for training high-fidelity world models that can serve as general-purpose simulators for policy evaluation and reinforcement learning.

\section*{Acknowledgments}
Apurva Badithela is supported by the Presidential Postdoctoral Research Fellowship at Princeton University. Samuel M. Bateman is supported by the National Science Foundation CISE Graduate
Fellowships under Grant No. 2313998. The authors were partially supported by Apple Inc. and the NSF CAREER Award \#2044149. Any views, opinions, findings, and conclusions or recommendations expressed in this material are those of the author(s) and should not be interpreted as reflecting the views, policies or position, either expressed or implied, of Apple Inc and the National Science Foundation.

\clearpage

\bibliographystyle{unsrtnat}
\bibliography{references}

\clearpage
\beginappendix{
\section{Implementation Details}
\label{appendix:implementation}

Here we provide more implementation details of \algname that expands upon Sec.~\ref{sec:method}, including the setup for play data collection (Sec.~\ref{sec:play data collection}), and training details for the action-conditioned video model (Sec.~\ref{sec:training}).

\smallskip
\noindent \textbf{Data Collection Design.} One ideal approach for data collection for training world models is to use an exploration policy that selects actions to maximally reduce the world model’s uncertainty (i.e., to intentionally seek out transitions that are most informative for the learned dynamics). While appealing in theory, we found this objective difficult to realize in a reliable, deployable system. In particular, using uncertainty estimates from a continually-updated video world model as a real-time reward signal can be unstable during robot execution, and can also create substantial operational overhead, since the model must be updated frequently as new data arrives, and the exploration signal can drift in ways that complicate monitoring and safety.

Instead of directly optimizing “informativeness” through world-model uncertainty, \algname uses a simple but robust proxy that maximizes the \emph{diversity} of behaviors while staying close to the policy’s natural action manifold. Empirically, this produces broad coverage of interaction modes and state transitions without requiring a brittle closed-loop interdependence between exploration and model uncertainty estimation. Concretely, to make the system readily portable across tasks with minimal engineering, we instantiate our play collector with $\pi_{0.5}$-DROID \cite{intelligence2025pi05visionlanguageactionmodelopenworld}. Beyond its general instruction-following capability, $\pi_{0.5}$-DROID reliably executes scene resets and handles a wide range of language prompts, which dramatically reduces human supervision, enabling persistent, scalable data collection. We provide further details analyzing the properties of the collected play data in Appendix~\ref{appendix:play_data}.

\begin{tcolorbox}[promptstyle_custom, title=Example Prompt]
    You are a robot that is trying to randomly manipulate/arrange objects on the table in a square region marked by tape.
    
    Ignore anything that's not on the gray table.
    
    First, observe if any object is outside of the square workspace area.
    
    If so, please output: {\color{blue!70} ``\verb|move the <> towards the center of the table."|}
    
    Otherwise, please come up with a short task for the robot to perform. 
    
    Possible instructions might include: 
    
    1) {\color{blue!70} ``\verb|Put the <> on the <>."|}
    
    2) {\color{blue!70} ``\verb|Remove the <> from the <>."|}
    
    3) {\color{blue!70} ``\verb|Put the <> near the <>."|}
    
    4) {\color{blue!70} ``\verb|Pick up the <> and put it onto the <>."|}
    
    Feel free to modify the prompt with different verbs and nouns (e.g. ``put" $\rightarrow$ ``move", ``block" $\rightarrow$ "object").
    
    Use color to help the robot better identify the objects. Don't include quotation marks in the task.
\end{tcolorbox}  

\begin{tcolorbox}[promptstyle_custom, title=Example In-Distribution Tasks]
    Put the carrot in the white bowl.
    
    Remove the white plush toy from the bowl
    
    Pick up the orange rectangular block and put it on the gray table.
    
    Fold the towel from left to right.
    ...
\end{tcolorbox}  
\begin{tcolorbox}[promptstyle_custom, title=Example Out-of-Distribution Tasks]
    Slide the carrot near the beige container.
    
    Move the orange rectangle onto the yellow cube.

    Flip the block up-side-down.
    
    Drag the blue towel across the table.
    ...
\end{tcolorbox}


\begin{wraptable}{r}{0.4\textwidth}
    \centering
    \caption{Hyperparameters used for video model training.}
    \label{tab:train-hparams}
    \begin{tabular}{@{}ll@{}}
    \toprule
    \textbf{Hyperparameter} & \textbf{Value} \\
    \midrule
    Learning Rate & 5e-6 \\
    Warmup Steps & 100 \\
    Scheduler & cosine \\
    Gradient clipping & 1.0 \\
    Global batch size & 64 \\
    \midrule
    Observation Shape & 320 $\times$ 192 $\times$ 3 \\
    Num. Predicted Frames & 5 \\
    Num. History Frames & 7 \\
    Frame-level Condition & True \\
    Frequency & 5 Hz \\
    Denoising Steps & 50 \\
    Guidance & 2.0 \\
    \bottomrule
    \end{tabular}
    \vspace{-8pt}
\end{wraptable}

\smallskip
\noindent \textbf{Video Model Training.} 
Our video model follows the setup of \cite{guo2025ctrl}, which jointly predicts 3 camera views (left, right, and wrist views, each with resolution 192 $\times$ 320) given the current view and the Cartesian-space end-effector pose. 
We first pre-train the base SVD model on the full DROID dataset, and then fine-tune using the curriculum described in Sec.~\ref{sec:training}, with the hyperparameters in Table~\ref{tab:train-hparams}. The goal of this curriculum is to progressively shift training mass toward harder, rarely occurring transitions, preventing the model from overfitting to dominant free-space motion patterns.
In practice, we compute a representation for each trajectory by concatenating CLIP embeddings from four sparsely-sampled frames over a 1-second window. For each task, we use $K=5$ to identify nominal behavior modes via clustering, and set $R=5$ ranks per task to construct the curriculum sampling buckets. We initialize the rank-sampling distribution as $(0.5,0.3,0.1,0.05,0.05)$ and anneal it to $(0.1,0.2,0.2,0.25,0.25)$ by uniformly updating the probabilities every $5000$ steps. Fig.~\ref{fig:kmeans} shows the most representative sample (centroid) from each cluster that is used to construct the ranks, illustrating that the discovered modes align well with intuitive task-level sub-behaviors.
While we used demonstration data as a proxy to define ranks with different motion attributes, we find that under this simplistic setup, the specific parameters used can have noticeable impact on model quality.
Future works could investigate how to make the curriculum-like setup more generalizable for data at larger scale, where a more robust signal for motion quality could potentially be directly learned from data.

\begin{figure*}
    \centering
    \includegraphics[width=\linewidth]{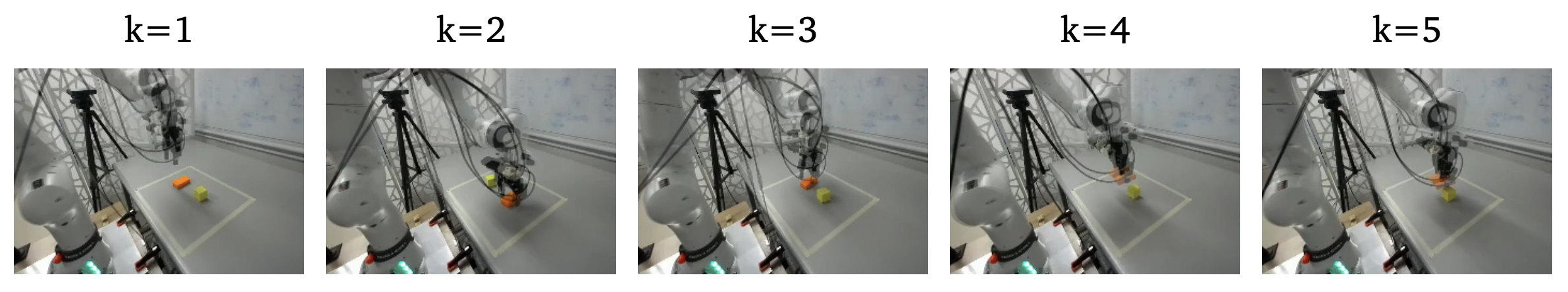}
    \vspace{-0.2em}
    \caption{Example extracted centroids from the demonstration data that roughly corresponds to distinct nominal behavior episodes.}
    \label{fig:kmeans}
    \vspace{-0.8em}
\end{figure*}
\section{Play Data Analysis}
\label{appendix:play_data}

\begin{figure*}[t]
    \centering
    \includegraphics[width=1.0\linewidth]{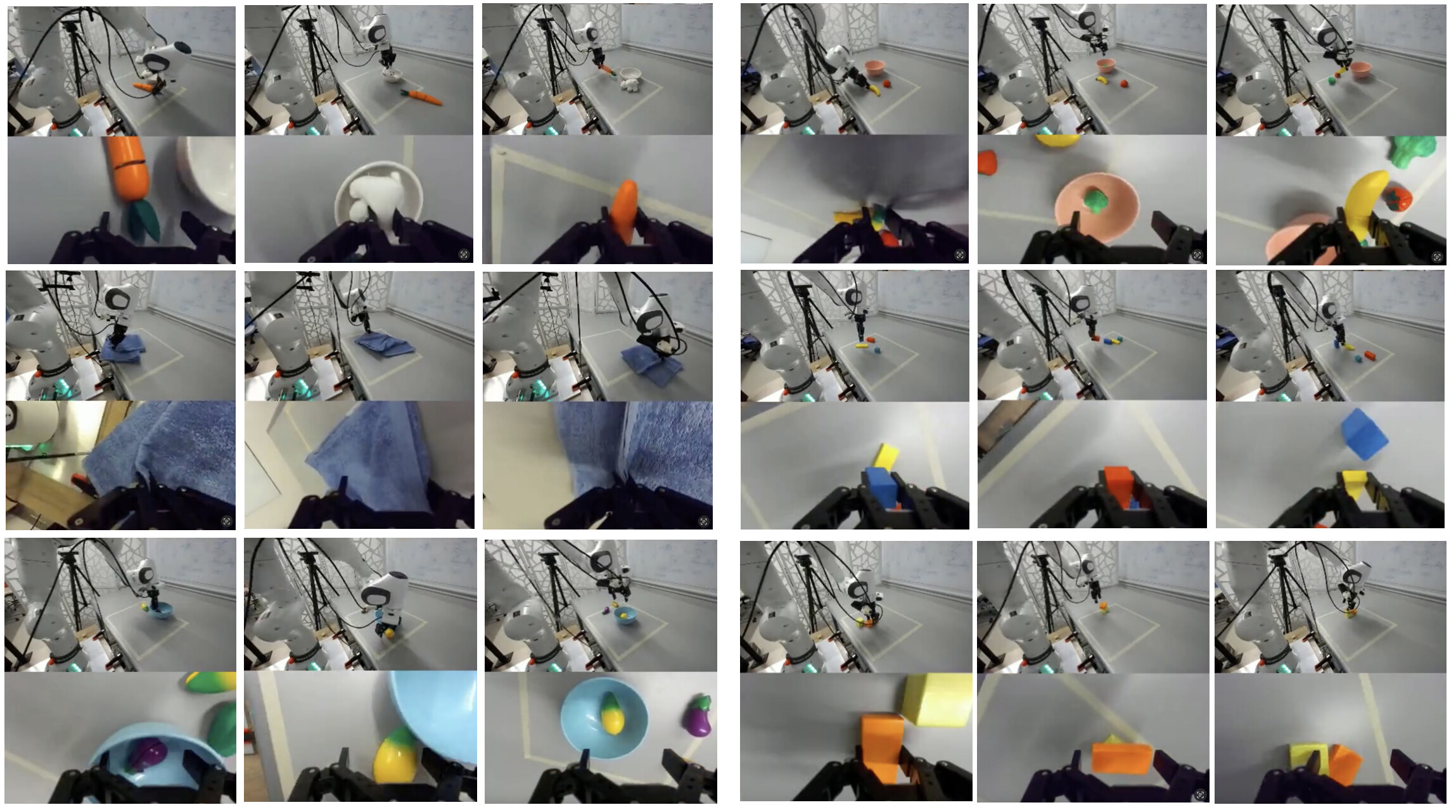}
    \caption{Example play data episodes selected from autonomous roll-outs for various tasks.}
    \label{fig:play_data_sample}
\end{figure*}

\begin{figure*}[!t]
    \centering
    \includegraphics[width=0.9\linewidth]{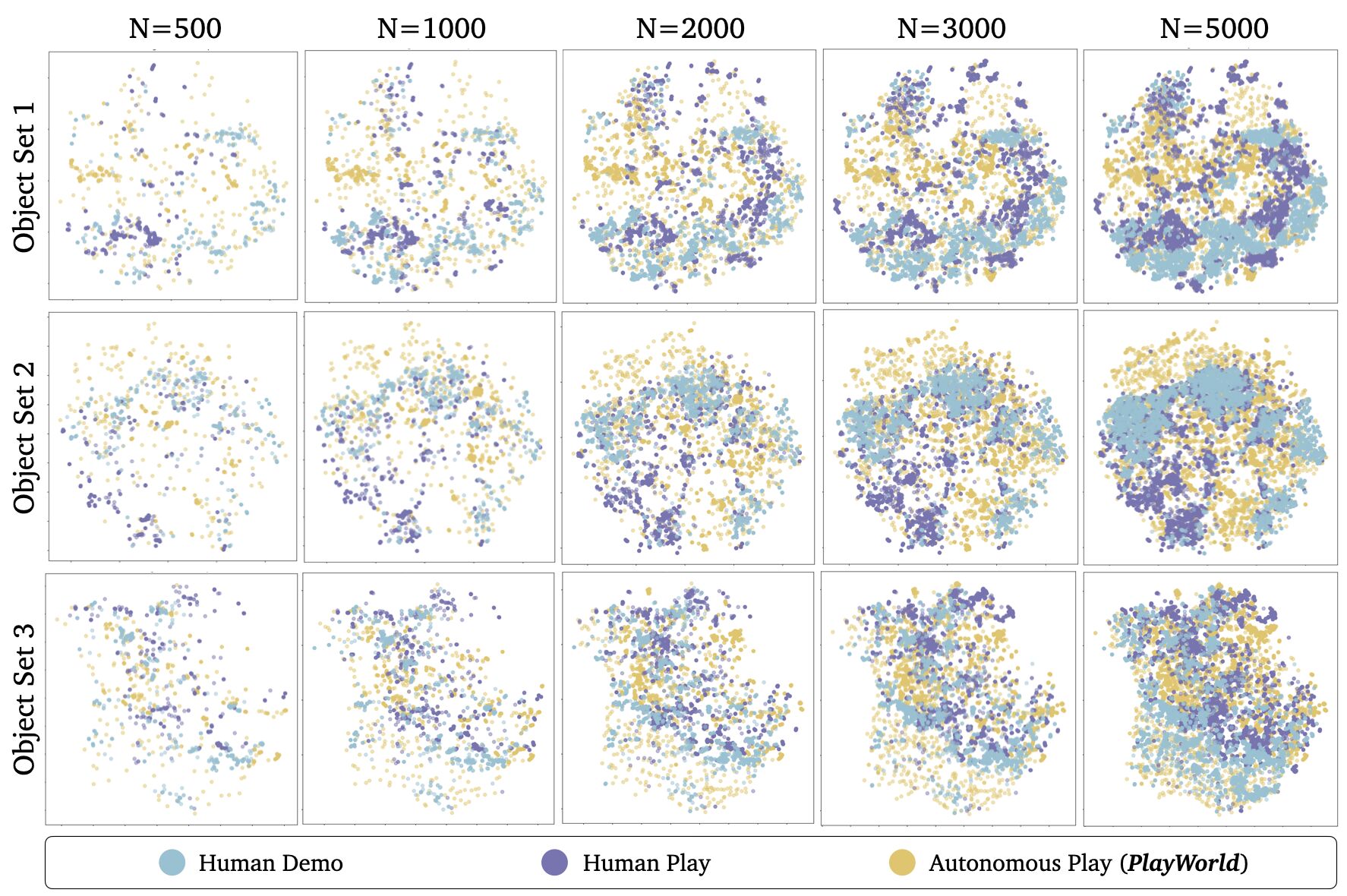}
    \caption{Training sample distribution for demonstration data, human-collected play data, and autonomously-collected play data at different scales. We observe that play data consistently provides the widest distribution over the state space, while other human-collected data exhibits limited expansion in coverage even at larger scale.}
    \label{fig:tsne_full}
\end{figure*}

Play data provides a scalable way to transform large volumes of a robot’s uncurated experience into a high-quality, simulatable environment, with minimal human involvement.
In this section, we provide further analysis on properties of play data in comparison to human-collected alternatives to demonstrate that autonomously-collected data can exhibit meaningful behavioral diversity.

\smallskip
\textbf{Play Data Visualization.}
Fig.~\ref{fig:play_data_sample} provides qualitative snapshots from the play data corpus on a few different object sets, highlighting the natural variations that emerge when the robot explores without a scripted task objective (e.g., diverse object contacts, viewpoints, and interaction outcomes). 

\smallskip
\textbf{Data Distribution at Different Scales.}
To complement these examples with a quantitative view, we include an extended t-SNE analysis (Fig.~\ref{fig:tsne_full}) across multiple dataset scales. Specifically, we embed the initial frame of each trajectory using CLIP features and visualize the resulting state-space coverage. Across matched data budgets, play data consistently spans a broader region of the embedding space than human-collected trajectories, while demonstration-based data concentrates in a comparatively narrow subset of states. As autonomously collected data scales beyond the practical limits of human collection, we expect it to cover the state space more densely and more broadly, yielding richer supervision and more informative transitions for training the world model.

\smallskip
\textbf{Concurrent Works.} Concurrent with our work, \cite{liang2026tetherautonomousfunctionalplay} explores the use of autonomous play for policy learning by generating task-directed interactions guided by vision-language models (VLMs). Their results show that play-collected data can achieve competitive performance compared to human demonstrations, highlighting the potential of autonomous play as a scalable and sustainable source of training data for robot learning.
\section{Additional Trajectory Replay Results}
\label{appendix:additional_results:replay}

In this section, we provide more details on the trajectory replay experiments, including the composition of our evaluation benchmark, details on the scaling experiments, and full results on additional perceptual metrics.

\smallskip
\noindent \textbf{Evaluation Benchmark.} 

\begin{wraptable}{r}{0.35\textwidth} 
\vspace{-0.5\baselineskip} 
\centering
\caption{Composition of the trajectory replay benchmark.}
\label{tab:replay_benchmark}

\parbox{0.35\textwidth}{%
\centering

\begin{subtable}[t]{\linewidth}
\centering
\begin{tabular}{@{}ll@{}}
\toprule
\textbf{Source} & \textbf{$\#$Samples} \\
\midrule
Human-collected & 97 \\
$\pi_0$ & 85 \\
$\pi_{0.5}$ & 122 \\
Diffusion Policies & 229 \\
\bottomrule
\end{tabular}
\end{subtable}

\vspace{0.6em}

\begin{subtable}[t]{\linewidth}
\centering
\begin{tabular}{@{}ll@{}}
\toprule
\textbf{Category} & \textbf{$\#$Samples} \\
\midrule
Success & 98 \\
Missed Grasp & 88 \\
Slide & 76 \\
Slip & 89 \\
Deformation & 90 \\
Collision & 92 \\
\bottomrule
\end{tabular}
\end{subtable}
} 
\vspace{-0.8\baselineskip} 
\end{wraptable}

Nominal task execution typically involves static contacts whose dynamics are comparatively consistent and easy to model. In contrast, dynamic object interactions such as collisions, deformation, or slip are substantially more diverse and harder to predict, yet they are often the decisive factors for counterfactual evaluation and task success. 
To target these failures, our benchmark is explicitly designed around predicting these interaction events.

Since contact events are highly multi-modal, we curate the benchmark from a large set of policy roll-outs and identify natural failure episodes that arise in practice. We then label each episode by failure mode, where the resulting distribution is summarized in Table~\ref{tab:replay_benchmark}. For every failure mode, our test set spans a wide range of trajectories. 
We argue that this coverage provides a more stringent and realistic assessment of action-conditioned video models, compared to typical nominal trajectory-replay evaluations which assess the model's capability of generating realistic trajectories, rather than faithfully predicting effects of actions. 
For completeness, we provide additional results on the perceptual metrics in Table~\ref{tab:main_results_extension} (extending the results in Table~\ref{tab:main_results}).


\begin{table*}[t]
    \centering
    \caption{Per-category perceptual similarity across different training data mixtures.}
    \label{tab:main_results_extension}

    \begin{subtable}{\textwidth}
        \centering
        \caption*{\textbf{PSNR} ($\uparrow$)}
        \begin{tabular}{lcccccc}
            \toprule
            Training Mix & Success & Missed Grasp & Slide & Slip & Deformation & Collision \\
            \midrule
           Human Demo & 24.78 & 24.37 & 23.98 & 23.66 & 22.58 & 24.67 \\
            Human Play & 23.56 & 24.41 & 23.90 & 24.01 & 22.93 &  24.91\\
            Robot Play & 24.95 & \cellcolor{mygreen!25}24.83 & 25.02 & \cellcolor{mygreen!25}24.84 & 23.22 & 24.61 \\
            Robot Play (Curriculum) & \cellcolor{mygreen!25}25.01 & 24.82 & \cellcolor{mygreen!25}25.14 & 24.78 & \cellcolor{mygreen!25}23.55 & \cellcolor{mygreen!25}24.95 \\
            \bottomrule
        \end{tabular}
    \end{subtable}
    
    \vspace{0.5em}

    \begin{subtable}{\textwidth}
        \centering
        \caption*{\textbf{MSE} ($\downarrow$)}
        \begin{tabular}{lcccccc}
            \toprule
            Training Mix & Success & Missed Grasp & Slide & Slip & Deformation & Collision \\
            \midrule
            Human Demo & 279.80 & 280.87 & 341.36 & 324.57 & 428.55 & 354.46 \\
            Human Play & 284.01 & 276.54 & 302.10 & 332.65 & 390.98 & 304.12 \\
            Robot Play & 274.81 & \cellcolor{mygreen!25}259.14 & 263.17 & 259.10 & 366.55 & 297.34 \\
            Robot Play (Curriculum) & \cellcolor{mygreen!25}265.13 & 264.19 & \cellcolor{mygreen!25}250.46 & \cellcolor{mygreen!25}243.51 & \cellcolor{mygreen!25}350.12 & \cellcolor{mygreen!25}295.53 \\
            \bottomrule
        \end{tabular}
    \end{subtable}
\end{table*}

\smallskip 
\noindent \textbf{Scaling Experiments.} In this section, we discuss the results in Fig.~\ref{fig:scaling} in greater detail.
To enable a fair and interpretable comparison across models trained with different dataset sizes, we convert the raw LPIPS metric $s_{raw}$ into a normalized score $s_{n}$ using a fixed affine transform: 
\begin{equation} 
s_{n} = -({s_{\text{raw}} - s_{\min}}) / ({s_{\max} - s_{\min}}).
\end{equation} 
Here, $s_{\min} = 0.072$ and $s_{\max} = 0.10$ are constants that define the LPIPS range used for normalization in our benchmark. This mapping preserves the ordering of models while (i) rescaling the narrow LPIPS interval to improve separation in scaling plots and (ii) inverting the axis so that higher values correspond to better perceptual fidelity.
As a result, scaling trends are easier to interpret, while all pairwise model comparisons remain unchanged.
\section{Additional Fine-Tuning Results}
\label{app:additional_finetuning}

Sim-to-real transfer is one of the most demanding end-to-end evaluations of a robotic simulator. Real robots expose compounded mismatch in contact, actuation, latency, and sensing, so policies trained in simulation often fail on hardware if 
they latch onto simulator-specific artifacts.
In this section, we show that an accurate world model can allow RL fine-tuning with consistent improvement in real-world success rate, even on challenging fine-grained manipulation tasks.

\smallskip
\noindent \textbf{Stable RL Fine-tuning for Diffusion-Based Models.}
Although many modern robot policies are parameterized as diffusion or flow-matching models, RL fine-tuning these generators remains challenging: the policy is realized through an iterative sampling procedure, making it difficult to obtain stable, low-variance gradients with standard actor--critic objectives.
Fine-tuning inside a learned world model introduces an additional source of brittleness, since effective exploration can hinge on the model capturing fine-grained action-dependent differences in predicted outcomes; small modeling errors can therefore distort the learning signal and lead to unstable updates.
As a result, prior works primarily use world models as data generators, performing supervised fine-tuning (SFT) on filtered synthetic rollouts instead of directly optimizing returns.
In \algname, we show that when the world model is sufficiently accurate, direct RL fine-tuning can be surprisingly effective even for fine-grained manipulation, and can improve robustness beyond what is achievable from demonstrations alone.

\smallskip
\noindent \textbf{Fine-Tuning with DSRL.}
For our fine-tuning setup, we adopt the formulation in \cite{wagenmaker2025steering}, which assumes a pretrained diffusion policy (e.g., trained with behavioral cloning) that generates an action sequence $\mathbf{a}$ conditioned on the current policy input $\mathbf{s}$ (observations and proprioceptive state).
Sampling proceeds by drawing an initial latent-noise vector $\mathbf{w}\sim\mathcal{N}(\mathbf{0},\mathbf{I})$ and running a fixed reverse-diffusion denoiser to obtain the action:
\begin{equation}
\mathbf{a} \;=\; g_{\theta}(\mathbf{s}, \mathbf{w}),
\end{equation}
where $g_{\theta}$ denotes the full reverse process (possibly multi-step) implemented by a frozen diffusion model with parameters $\theta$.
This induces the (marginal) policy distribution
\begin{equation}
\pi_{\theta}(\mathbf{a}\mid \mathbf{s})
\;=\;
\int \delta\!\left(\mathbf{a}-g_{\theta}(\mathbf{s},\mathbf{w})\right)\,\mathcal{N}(\mathbf{w};\mathbf{0},\mathbf{I})\,d\mathbf{w},
\end{equation}
i.e., actions are generated by pushing forward a simple noise prior through the denoiser.

Following the ``noise actor'' formulation of \citet{wagenmaker2025steering}, we replace the fixed prior $\mathcal{N}(\mathbf{0},\mathbf{I})$ with a learned, state-conditioned distribution over initial noise:
\begin{equation}
\mathbf{w}\sim \pi_{\phi}^{\mathcal{W}}(\mathbf{w}\mid \mathbf{s}),
\qquad
\mathbf{a}=g_{\theta}(\mathbf{s},\mathbf{w}),
\end{equation}
where $\pi_{\phi}^{\mathcal{W}}$ is a small policy (e.g., an MLP Gaussian) with parameters $\phi$.
This yields a \emph{steered} action policy
\begin{equation}
\pi_{\phi,\theta}(\mathbf{a}\mid \mathbf{s})
\;=\;
\int \delta\!\left(\mathbf{a}-g_{\theta}(\mathbf{s},\mathbf{w})\right)\,\pi_{\phi}^{\mathcal{W}}(\mathbf{w}\mid \mathbf{s})\,d\mathbf{w},
\end{equation}
which can be viewed as learning an expressive adapter in the diffusion model's latent-noise space while leaving the diffusion parameters $\theta$ unchanged.
We optimize $\pi_{\phi}^{\mathcal{W}}$ with an RL objective defined on environment rollouts produced by executing $\mathbf{a}=g_{\theta}(\mathbf{s},\mathbf{w})$.
Let $\tau=(\mathbf{s}_0,\mathbf{w}_0,\mathbf{r}_0,\mathbf{s}_1,\dots)$ denote the resulting trajectory; then the goal is
\begin{equation}
\max_{\phi}\;
J(\phi)
\;=\;
\mathbb{E}_{\tau\sim(\pi_{\phi}^{\mathcal{W}},\,g_{\theta})}
\left[
\sum_{t\ge 0}\gamma^t\,\mathbf{r}_t
\right].
\end{equation}
Crucially, we treat $\mathbf{w}$ as the RL action and the (frozen) diffusion sampler $g_{\theta}$ as part of the environment interface. In practice, this allows standard off-policy actor--critic updates using replay tuples
$(\mathbf{s}_t,\mathbf{w}_t,\mathbf{r}_t,\mathbf{s}_{t+1})$,
without requiring backpropagation through the world dynamics or through the multi-step reverse-diffusion procedure. The specific hyperparameters used can be found in Table~\ref{tab:dsrlna-hparams}.

\smallskip
\noindent \textbf{Reward Model Design.}
To enable efficient policy improvement, we train a simple progress-based reward model from a small set of demonstrations.
Given two time steps, the model predicts task progress, which we convert into a dense per-step signal by taking the difference in predicted progress between consecutive steps:
\[
r_t \;=\; \hat{p}(s_{t+1}) - \hat{p}(s_t).
\]
Despite its simplicity, this reward signal leads to fast and stable learning in our setting: we often observe measurable behavioral improvements within the first 10--20 episodes.
In contrast, sparse terminal $0/1$ rewards require substantially more interaction to achieve comparable gains.
We also explored vision-language-model (VLM) rewards, but found them less reliable in practice, often destabilizing fine-tuning with false predictions.

\begin{table}[!htbp]
\centering
\caption{Hyperparameters used for DSRL training and diffusion policy.}
\label{tab:dsrlna-hparams}
\begin{tabular}{@{}ll@{}}
\toprule
\textbf{Hyperparameter} & \textbf{Value} \\
\midrule
Obs dim & 8 \\
Action dim & 8 \\
Image shape & [6, 192, 192] \\
Cond steps (state) & 1 \\
Cond steps (image) & 1 \\
Horizon & 16 \\
Max episode steps & 30 \\
\midrule
Actor LR & 1e-5 \\
Batch size & 64 \\
Discount & 0.99 \\
Train freq & 15 \\
UTD & 10 \\
Tau & 0.001 \\
Layer norm & True \\
MLP backbone & 3 layers $\times$ 256 \\
Target Entropy & 0.0 \\
Initial rollout steps & 800 \\
Action magnitude & 0.5 \\
\midrule
Denoising steps & 100 \\
DDIM steps & 5 \\
Prediction target & $\epsilon$ \\
Spatial embedding & 128 \\
\bottomrule
\end{tabular}
\end{table}

\smallskip
\noindent \textbf{Experiment Setup.}
To show the power of fine-tuning with \algname, we use diffusion policies trained from scratch with $5-10$ demonstrations. We then collect $10-20$ initial object configurations as environment initializations for the world model, and perform fine-tuning by randomly sampling among the available initializations. With each checkpoint, we perform $20$ hardware trials to calculate the success rates shown in Fig.~\ref{fig:finetune}.

\smallskip
\noindent \textbf{Effectiveness of Fine-Tuning and World Model Hacking.}
Here, we report some interesting observations from the fine-tuning experiments. 
First, we find that \emph{stochastic predictions from the world model are beneficial for learning robust policies}, particularly under out-of-distribution (OOD) conditions.
We hypothesize that stochastic video rollouts effectively act as a form of data augmentation: by exposing the policy to diverse, plausible futures during training, the resulting behavior generalizes to initializations that were not observed during either pretraining or fine-tuning.
As shown in Fig.~\ref{fig:towel_ft}, when initialized from a novel configuration, the fine-tuned policy is able to gradually adapt its actions and eventually achieve a successful towel grasp.

Second, we find that \emph{world-model hallucinations are not necessarily as detrimental to fine-tuning as one might expect}.
In several baseline fine-tuning runs, the learned policy partially collapsed into exploiting spurious model predictions---executing actions that are suboptimal (or nonsensical) in reality, yet yield high reward under the learned dynamics (i.e., ``reward hacking'' within the model).
Despite this behavior in imagination, the resulting policies often still transferred well to the real robot.
We hypothesize that fine-tuning can simultaneously (i) extract genuinely useful signal from the world model and reward, but may (ii) overfit to hallucinated failure modes that are unlikely to occur under real dynamics; when deployed, these spurious gradients may have limited effect because the corresponding states are never realized.
Overall, this suggests that meaningful performance gains are possible even with imperfect world models, provided that hallucinations do not dominate the policy updates that matter on real trajectories.

\begin{figure*}
    \centering
    \includegraphics[width=\linewidth]{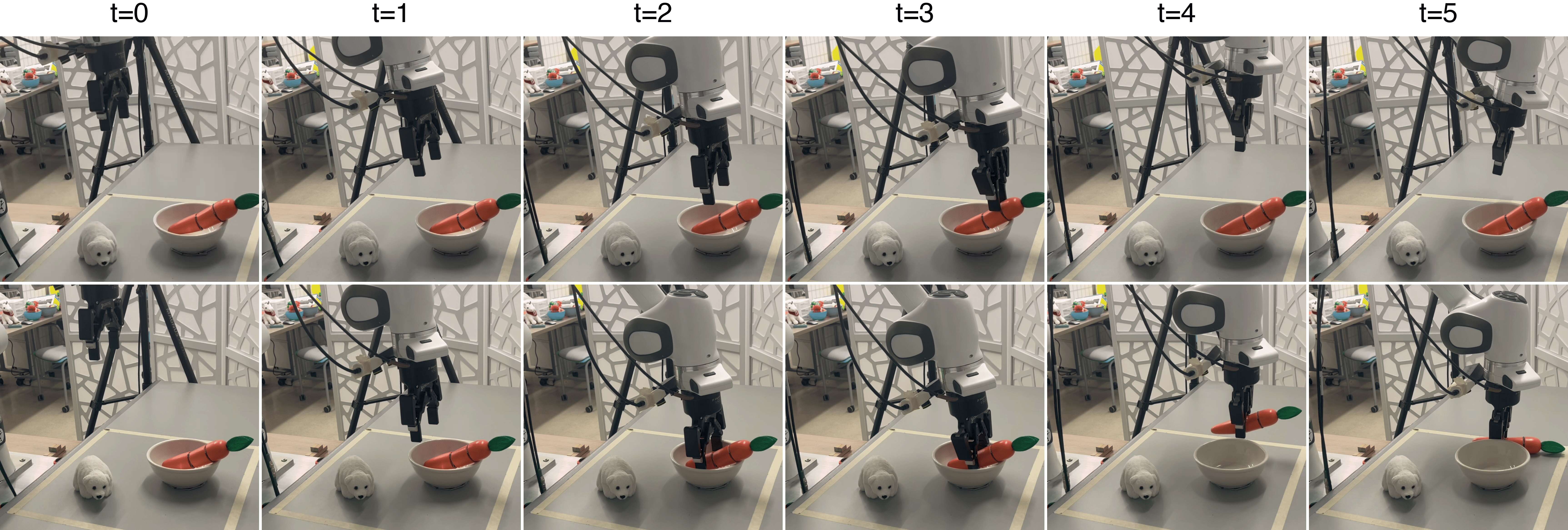}
    \caption{Task: remove the carrot from the bowl. Top: base policy behavior in the real world. Bottom: \algname fine-tuned policy behavior in the world, where the robot learned to grasp a more stable position.}
    \label{fig:carrot_ft}
\end{figure*}

\begin{figure*}
    \centering
    \includegraphics[width=\linewidth]{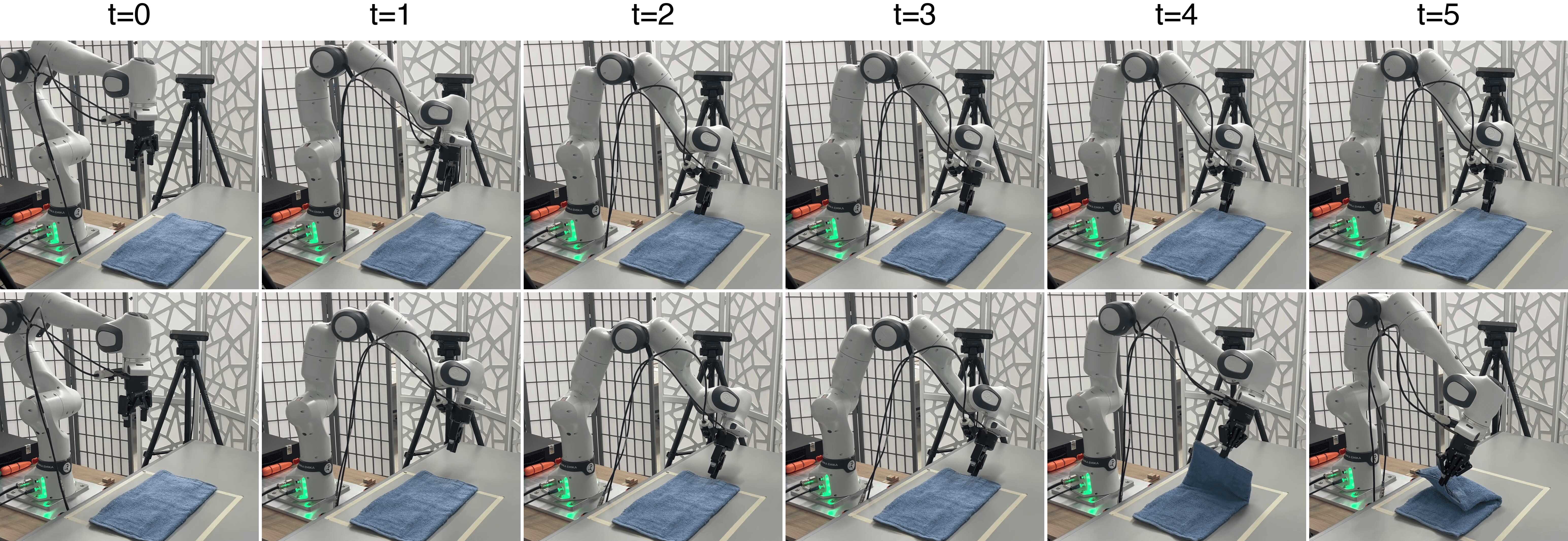}
    \caption{Task: fold the towel from left to right. Top: base policy behavior in the real world. Bottom: \algname fine-tuned policy behavior in the world, where the grasping motion is more robust and accurate.}
    \label{fig:towel_ft}
\end{figure*}

Finally, we also observe the converse phenomenon: policies can \emph{overfit to the world model} and degrade with prolonged fine-tuning.
Because even small prediction errors (e.g., in depth, contacts, or control response) can shift the boundary between ``success'' and ``failure'' in the model, the set of trajectories that are scored as successful in imagination does not perfectly align with real-world outcomes.
As a result, extended fine-tuning can gradually bias the policy toward exploiting these mismatches, yielding apparent gains under the model while reducing real-world success rate.
This effect is most pronounced when the policy repeatedly revisits a narrow subset of imagined high-reward states, effectively specializing to idiosyncrasies of the learned dynamics rather than improving the underlying task strategy.
Practically, this motivates using conservative stopping criteria (e.g., early stopping on real rollouts) and/or regularization toward the pretrained policy to prevent overfitting to model errors.

\smallskip
\noindent \textbf{Related/Concurrent Works.}
Prior work on adapting video-based world models for RL-finetuning largely falls into two categories. The first trains \textbf{task-specific world models}, fine-tuned on narrow, single-task datasets to optimize performance for a predefined objective \cite{Chandra2025DiWADPA, zhu2025wmpo}. While effective within scope, these models often overfit to a limited interaction distribution and lack generality beyond the target task. The second direction explores \textbf{generalist world model fine-tuning}, where large pretrained video models are adapted using broad, internet-scale or multi-task data \cite{sharma2026worldgymnasttrainingrobotsreinforcement}. These approaches typically improve high-level capabilities such as language following, semantic consistency, or robustness to distractors, but do not explicitly optimize for contact-rich physical fidelity required for fine-grained visuo-motor control.
\algname instead explores an intermediate space by collecting lab-scale interaction data that spans a broad distribution of contact events beyond any single task. We show that the resulting world model more faithfully captures fine-grained object motion and contact dynamics, enabling policy improvements that transfer reliably to the real world—yielding smoother execution, more accurate and robust end-effector control, and higher success rates on fine-grained manipulation tasks.

\section{Final Remarks}
We believe large-scale play data is a particularly promising and scalable supervision source for learning video world models, which can make learned simulators substantially more practical for policy development.
In our lab setting, we were able to collect many hours of diverse, unlabeled interaction data with relatively low overhead, demonstrating its utility for various post-training applications.
Looking forward, scaling data collection across multiple institutions and hardware setups could further expand coverage of environments, object sets, and embodiments, improving both the fidelity and the robustness of learned dynamics.
Such a shared play-data ecosystem would enable us to build stronger, more generalizable world simulators that will be a valuable resource to the robot learning community.



}

\end{document}